\documentclass[letterpaper, 10 pt, conference]{ieeeconf}  % Comment this line out if you need a4paper

\IEEEoverridecommandlockouts                              % This command is only needed if 
\usepackage{graphics} % for pdf, bitmapped graphics files
\usepackage{epsfig} % for postscript graphics files
\usepackage{times} % assumes new font selection scheme installed
\usepackage{amsmath} % assumes amsmath package installed
\usepackage{amssymb}  % assumes amsmath package installed
\usepackage{color} % change text color 
\usepackage{booktabs}
\usepackage{subcaption}
\usepackage{algorithm}
\usepackage{algorithmic}
\usepackage{calc}
\usepackage{multirow}
\title{\LARGE \bf
Context-Aware Deep Lagrangian Networks for Model Predictive Control
}

\author{
  Lucas Schulze$^{1}$,
  Jan Peters$^{1,2,3,4}$,
  Oleg Arenz$^{1}$% <-this % stops a space
\thanks{
This work was funded by the German Research Foundation (DFG) - Project number
PE 2315/18-1, and the German Federal Ministry of Re-
search, Technology and Space (BMFTR) - Project number
01IS23057B. This project has been supported by a hardware
donation by NVIDIA through the Academic Grant Program.\newline$^{1}$Department of Computer Science, Technical University of Darmstadt, Germany.
$^{2}$Hessian.AI.
$^{3}$German Research Center for AI (DFKI), Research Department: Systems AI for Robot Learning.
$^{4}$Robotics Institute Germany (RIG).
\newline Corresponding author: {\tt\small lucas.schulze@tu-darmstadt.de }
}}

\newcommand{\Transp}{^\mathrm{\scriptscriptstyle T}}

\newcommand{\ZeroMat}{\mathbf{0}}

\newcommand{\jointTau}{\boldsymbol{\tau}}
\newcommand{\jointPos}{\mathbf{q}}
\newcommand{\jointVel}{\dot{\mathbf{q}}}
\newcommand{\jointAcc}{\ddot{\mathbf{q}}}

\newcommand{\jointPosErrNorm}{\mathrm{e}}
\newcommand{\jointVelErrNorm}{\dot{\mathrm{e}}}

\newcommand{\jointPosInit}{\mathbf{q}_0}

\newcommand{\jointPosRef}{\mathbf{q}_\mathrm{d}}
\newcommand{\jointVelRef}{\dot{\mathbf{q}}_\mathrm{d}}

\newcommand{\PDKp}{\mathbf{K}_\mathrm{p}}
\newcommand{\PDKd}{\mathbf{K}_\mathrm{d}}

\newcommand{\Qmpc}{\mathbf{Q}}
\newcommand{\Rmpc}{\mathbf{R}}
\newcommand{\fmodel}{f}

\newcommand{\Qlqr}{\mathbf{Q}_\mathrm{LQR}}
\newcommand{\Rlqr}{\mathbf{R}_\mathrm{LQR}}

\newcommand{\xss}{{\mathbf{x}}}
\newcommand{\uss}{{\mathbf{u}}}
\newcommand{\yss}{{\mathbf{y}}}

\newcommand{\Ny}{\mathrm{n_y}}

\newcommand{\xref}{\xss^\mathrm{d}}
\newcommand{\Xopt}{\bar{\xss}}
\newcommand{\Uopt}{\bar{\uss}}
\newcommand{\hConst}{\mathbf{h}}

\newcommand{\EKin}{K}
\newcommand{\EPot}{P}
\newcommand{\inertiaMat}{\mathbf{H}}
\newcommand{\inertiaTri}{\mathbf{L}}
\newcommand{\inertiaTriDiag}{\mathbf{l}_{\mathrm{diag}}}
\newcommand{\inertiaTriOff}{\mathbf{l}_{\mathrm{off}}}
\newcommand{\paramInertia}{\boldsymbol{\theta}_\inertiaMat}
\newcommand{\paramPot}{\boldsymbol{\theta}_\EPot}

\newcommand{\latentVariableZ}{\mathbf{z}}

\newcommand{\EoMC}{\mathbf{C}}
\newcommand{\EoMG}{\mathbf{g}}

\newcommand{\histInput}{\mathbf{h}}

\newcommand{\histLength}{\mathrm{n}_\mathrm{h}}

\newcommand{\inertiaMatNom}{\hat{\mathbf{H}}}
\newcommand{\EoMCNom}{\hat{\mathbf{C}}}
\newcommand{\EoMGNom}{\hat{\mathbf{g}}}
\newcommand{\jointTauNom}{\hat{\boldsymbol{\tau}}}

\newcommand{\jointTauDiffNom}{\tilde{\boldsymbol{\tau}}}
\newcommand{\inertiaMatDiff}{\tilde{\mathbf{H}}}

\newcommand{\EoMGDiff}{\tilde{\mathbf{g}}}

\newcommand{\chirpA}{\mathbf{a}_\mathrm{e}}
\newcommand{\chirpB}{\mathbf{b}_\mathrm{e}}
\newcommand{\chirpOrder}{\mathrm{n}_\mathrm{e}}
\newcommand{\chirpPeriod}{\mathrm{T}_\mathrm{e}}
\newcommand{\chirpOmega}{\omega_\mathrm{e}}

\newcommand{\KFQ}{\mathbf{Q}_\mathrm{EKF}}
\newcommand{\KFR}{\mathbf{R}_\mathrm{EKF}}

\newcommand{\pee}{\mathbf{p}}

\newcommand{\Jee}{\mathbf{J}_\mathrm{ee}}
\newcommand{\Fee}{\mathbf{f}_\mathrm{ee}}
\newcommand{\dotFee}{\dot{\mathbf{f}}_\mathrm{ee}}

\newcommand{\EKFFeeEst}{\hat{\mathbf{f}}_{ee}}
\newcommand{\EKFTaueeEst}{\hat{\boldsymbol{\tau}}_\mathrm{ee}}

\begin{document}

\maketitle
\thispagestyle{empty}
\pagestyle{empty}

%% TODO:
% - Acknowledgment Melanie
% - Improve figures
% - Add denomination: white-box, grey-box and black-box methods
% - Improve equations
% - Add reference universal policy

%%%%%%%%%%%%%%%%%%%%%%%%%%%%%%%%%%%%%%%%%%%%%%%%%%%%%%%%%%%%%%%%%%%%%%%%%%%%%%%%
\begin{abstract}

Controlling a robot based on physics-consistent dynamic models, such as Deep Lagrangian Networks (DeLaN), can improve the generalizability and interpretability of the resulting behavior. However, in complex environments, the number of objects to potentially interact with is vast, and their physical properties are often uncertain. This complexity makes it infeasible to employ a single global model. Therefore, we need to resort to online system identification of context-aware models that capture only the currently relevant aspects of the environment. While physical principles such as the conservation of energy may not hold across varying contexts, ensuring physical plausibility for any individual context-aware model can still be highly desirable, particularly when using it for receding horizon control methods such as model predictive control (MPC). Hence, in this work, we extend DeLaN to make it context-aware, combine it with a recurrent network for online system identification, and integrate it with an MPC for adaptive, physics-consistent control. We also combine DeLaN with a residual dynamics model to leverage the fact that a nominal model of the robot is typically available. We evaluate our method on a 7-DOF robot arm for trajectory tracking under varying loads. Our method reduces the end-effector tracking error by 39\%, compared to a 21\% improvement achieved by a baseline that uses an extended Kalman filter. 

% https://ieeexplore.ieee.org/stamp/stamp.jsp?arnumber=8909368
% https://ieeexplore.ieee.org/stamp/stamp.jsp?arnumber=8968268%
% https://arxiv.org/pdf/2202.09834

% Kinematic assumptions, no agile motions
\end{abstract}

%%%%%%%%%%%%%%%%%%%%%%%%%%%%%%%%%%%%%%%%%%%%%%%%%%%%%%%%%%%%%%%%%%%%%%%%%%%%%%%%
\section{Introduction}
% What's the history of the paper? Focus on manipulation?
% Physically consistent online system identification for model predictive control

Representative models are essential for control of dynamic
systems, whether for learning approaches, such as
reinforcement or imitation learning in simulation,
or for model-based techniques, like model predictive control
(MPC).
White-box models can be derived from first principles, but they often require significant engineering effort and may not generalize well to dynamically changing environments.
To address these challenges, online system identification (SysID) continuously adapts the model using online observations, improving its accuracy and robustness during operation.

Classical adaptive control techniques online identify the model's parameters based on a linear regressor~\cite{slotine_regressor,est_inertia_an,friction_adaptive, nadia}. However, these methods rely on prior knowledge, such as a known kinematic tree, and assume persistent excitation \cite{model_learning_control}. Additionally, they are typically designed for specific types of model mismatches, limiting their generalization across different systems and conditions.

On the other hand, black-box approaches leverage recorded data to model the environment using nonparametric models, such as Gaussian process regression (GPR)~\cite{gpr_tuong, gp_mpc,safe_gpr_mpc} or locally weighted projection regression~\cite{load_estimation_lwpr}, or parametric models such as neural networks~\cite{nmpc_rbf}. However, these techniques typically perform well only within the trained domain and require large amounts of data.

A powerful technique to increase the accuracy of the learned model in the small data regime, is to employ physics-informed inductive biases~\cite{watson2024}.
To address physical plausibility in offline SysID, grey-box methods such as Deep Lagrangian Networks (DeLaN) were proposed to combine Lagrangian mechanics with deep learning \cite{delan_lutter2018}, \cite{delan_2021}. The resulting model from DeLaN is interpretable and can be used as a forward and inverse model. A similar approach is used in Hamiltonian Neural Networks (HNN)~\cite{HNN}, which are based on Hamiltonian mechanics. Both DeLaN and HNN can be applied to model-based control~\cite{delan_4ec, neural_ode_lie_control}.

%%%To address physically plausible in offline SysID, Deep Lagrangian Networks (DeLaN) and Hamiltonian Neural Networks (HNN) were proposed to combine Lagrangian and Hamiltonian mechanics with deep learning \cite{delan_2021}. The resulting model from DeLaN is interpretable and can be used as a forward and inverse model, which can be used for model-based control~\cite{delan_4ec}. 
%Using a nominal model as an inductive bias focuses the learn-ing process on the unknown components, as some robot components remain constant over time and across different environments, while we can still use available robot models. Additionally, learning residual functions is more efficient and requires less data with deep neural network

However, these offline methods are not suitable for environments where the robot manipulates different objects, as they cannot account for the resulting changes in the robot's dynamics.
Therefore, we propose an extension of DeLaN to enable physically consistent online SysID by learning a contextual DeLaN model that obtains a latent representation of the environment as additional input, which can be identified online.
This modification allows us to generalize across multiple and time-varying environments while maintaining a physically plausible model at any given time. By using a fixed model during each MPC iteration, which might be updated between iterations, we impose the prior belief that the dynamic properties will remain constant over the prediction horizon.
Yet, this approach still enables swift adaptation whenever this prior belief is violated. We argue that this method is particularly fruitful in settings where changes in dynamics are sparse over time, e.g., when picking up or dropping objects.
%
%Our motivation to use a nominal model as an inductive bias is to focus the learning on the unknown components since some robot components remain constant over different environments.
%By using a nominal model as an inductive bias, we focus on learning the unknown component, as some robot components remain constant over time and across different environments, while we can still use available robot models.
%

Furthermore, we use a nominal model as an inductive bias to focus on learning the unknown dynamics of the system, as certain elements may remain unchanged compared to the nominal case. Learning a residual dynamics model that incorporates prior dynamics has been shown to improve data efficiency and generalization \cite{model_knowledge_learn}. In the context of deep learning, learning a residual model has similar benefits \cite{residual_deep_residual_learning}.
%
%\cite{model_knowledge_learn} demonstrates the benefit of using prior model knowledge in terms of data-efficient and generalization of the learned model. In the case of deep learning, learning a residual model is also more data-efficient \cite{residual_deep_residual_learning}. 
%
It also enables the use of smaller networks, which is essential for fast evaluation and optimization, which is a key requirement of MPC.

%Our motivation for learning a residual model is easier to learn residual functions \cite{residual_deep_residual_learning}, and we can explore nominal models. Furthermore, physic-informed networks require fewer data and extrapolate well out of the training domain, which are desirable characteristics for online SysID.

%\replaced{We}{To} condition our model on real-time data, \replaced{by using a history encoder based on}{our method has a temporal encoder using} a long short-term memory (LSTM) network\deleted{ as a hypernetwork~\cite{hypernetworks_review}}. Thus, we obtain a latent environment description based on \replaced{historical}{recent} joint positions, velocities, and residual torques from the nominal model. This latent variable is then included in the original DeLaN networks as an additional input.

%Furthermore, we leverage DeLaN’s feature to provide both forward and inverse dynamics models by integrating our \highlight{Residual DeLaN} into an MPC. This combination enables the controller to solve constrained optimal problems while reasoning unknown dynamics, which benefits agile motions and complex robotic tasks.

Our work is within the context of meta-learning, i.e., learning how to optimize a meta-objective over different tasks using previous experience~\cite{meta_learning_nn_survey}. 
Many approaches combine this concept with adaptive control, relying on GPR~\cite{gpr_hyperparamater, Arcari2020MetaLM, gpr_mpc_arm}, non-linear basis functions~\cite{Richards2021AdaptiveControlOrientedMF,bayesian_learning_mpc}, or feed-forward neural networks~\cite{Lapandi2024MetaLearningAM_org,up_rss_2017}.
%
%However, these methods do not ensure physical consistency. \added{
%Our work is also related to \cite{belbute2021hyperpinn} and \cite{HyperLRPINNs_2023}, where a hypernetwork is used to estimate a latent input to a physics-informed neural network.}
%\added{While these methods do not ensure physical consistency, similar meta-learning methods have been applied to physics-informed neural networks (PINNs)~\cite{pinn}. For example, iMODE~\cite{Li2023} learns a more general PINN that incorporates an additional latent variable to describe the concrete system, similar to our approach.}
%
Similar meta-learning methods have been applied to physics-informed neural networks (PINNs)~\cite{pinn}, iMODE~\cite{Li2023} learns a more general PINN that incorporates an additional latent variable to describe the underlying system, similar to our approach. However, these methods do not guarantee physical consistency.

%Our work is in the context of using meta-learning, or learning to learn as defined by (cite survey in bayesian). These methods rely on nonparametrics techniques: Gaussian Processes regression (), sinusoidal~\cite{bayesian_learning_mpc} or non-linear basis functions~\cite{Richards2021AdaptiveControlOrientedMF}. However, these are black-box methods and don't offer physical plausiblity.

Similarly, while residual models have been previously employed for MPC \cite{knode_mpc, l4_casadi_taylor_salzmann2023neural}, these methods did not use physically plausible models for better data efficiency.
Furthermore, while \cite{Duong2021AdaptiveCO} proposes an adaptive controller based on Hamiltonian neural ordinary differential equations to learn disturbance features, ensuring physical consistency, their method is only applied to a feedback energy control strategy, and the results are demonstrated only in simulation.

%Furthermore, we leverage DeLaN’s feature to provide both forward and inverse dynamics models by integrating our Residual DeLaN into an MPC. This integration enables the controller to reason about unknown dynamics, which is beneficial for agile motions and complex robotic tasks.

%We demonstrate the generalization capability of our method by doing zero-shot experiments in the real manipulator using a model trained only with simulated data. As well we demonstrated it's benefits in terms of joint trajectory tracking.
We demonstrate the adaptability of our method in joint trajectory tracking of a 7-DOF Franka Emika Panda under an unknown payload, both in simulation and on the real robot. 
To demonstrate the benefits of identifying an entire dynamics model and not only gravitational compensation for agile motions, we compare it to an extended Kalman filter (EKF) that directly estimates the force at the end-effector. As shown in our experiments, we outperform the tracking and prediction error of both, the nominal model and the EKF. On the real robot, our method is applied zero-shot, that is,  we apply it using a model trained solely with simulated data.

%We take advatange of physics simulator to collect data, and perform esperiments in simulation and in the real robot, demonstraing the perfomance of our method.

%In offline SysID, phydeep lagrangian networks and hamiltonian were proposed for physical consistent and plausible. In the context of physically plausible learned methods, deep Lagrangian networks \cite{delan_2021} was proposed to learn a model that obey Lagrangian Mechanics. In \cite{delan_4ec}, the authors proposed an energy based controller using a learned model

%An interesting feature of DeLaN is the possibility to do both forward and inverse dynamics from the learned model. More specific, a function is obtained from the network parameters. 

%Based on this feature, we propose to use a MPC using a learned model. Therefore, the MPC can \textit{reason} about the effect of the unkwown dynamics, this is specially important for agile motions.

%GPR also offer this posibility, as done~\cite{gp_mpc}. However, GPR suffers with the curse of dimensionality, and are ussualy aplied to a small number of variables, as each output needs to be a single GPR. Also, GPR are not physical consistent. 
%Alternative learning-based methods, such as Gaussian Process Regression (GPR), have also been used for MPC~\cite{gp_mpc}. However, GPR suffers from the curse of dimensionality, making it impractical for high-dimensional systems, as a separate Gaussian Process must be trained for each output variable. Additionally, GPR lacks physical consistency, as it does not inherently enforce constraints like energy conservation.

\subsection{Contributions}
Our main contributions are as follows:
\begin{itemize}
    \item We combine DeLaN with a nominal model to only predict the unmodeled dynamics.
    \item We extend DeLaN to a contextual setting, enabling us to learn a single network for different dynamics.
    \item By training our contextual DeLaN along with a history encoder in simulation, we are able to online identify a physically plausible dynamics model.
    \item We combine the aforementioned contributions for MPC by using recent software frameworks~\cite{acados_2021,HPIPM,carpentier2019pinocchio,l4_casadi_taylor_salzmann2023neural} to integrate our contextualized DeLaN network with an optimizer suitable for real-time optimization.
    \item We demonstrate the benefits of our learning-based control method for agile manipulation tasks, by performing zero-shot real robot experiments from simulated data.
\end{itemize}

%The remainder of this paper is organized as follows: Section~\ref{sec:res_delan} briefly introduces the concept of Deep Lagrangian Networks\added{ and presents our extensions for the residual and contextual setting as well as their application for MPC. Section~\ref{sec:experiments} presents our experimental setting as well as the results. Finally, Section~\ref{sec:conclusion} summarizes our findings. }  

The remainder of this paper is organized as follows: Section~\ref{sec:res_delan} briefly reviews DeLaN and presents our extensions for the residual and contextual setting, along with their application for MPC: Context-Aware Deep Lagrangian MPC (CaDeLaC). Section~\ref{sec:experiments} presents our experimental setting as well as the results. Finally, Section~\ref{sec:conclusion} summarizes our findings.

% Therefore, our paper has the following contributions:
% \begin{itemize}
%     \item Extension of Delan to Multi Environments
%     \item Integration of Delan with MPC
%     \item Zero-shot experiments from simulated data
%     \item Have learned model is a function useful for MPC
% \end{itemize}
% % MPC
% Pros
% - Ensure constraints
% - Physical consistent
% - Easily to use known knowledge

% Cons
% - Computational cost
% - Model quality: not everything is modelable

% MPC + System identifcation (SI)
% Pros
% - Requires Low data
% - Generalize well

% Cons
% - Rely on some format of the model mismatch: kinematic structur -> tailor made for specific problems
% - Model mismatch as a constant disturbance

% MPC + Learning
% Cite some approaches: Model based RL, learn model used inside MPC
% Pros
% - In the Model based RL, helps to use low data, since it combines the optimization over the learned model

% Pure Learning
% - Too much data

% Delan
% Pros
% - Physical consistent model. Few assumptions about the mode (no kinematic structure)
% - Not explored for control

% Baselines
% - SI with kalman filter to identify the model mismatch
% - RL with everything
% - Supervised learning to learn the model

% General problem:
% The MPC can reasonable how the change of the variables affect the disturbance, non-stationary disturbance? 

% Contributions
% Flexible framework which can account for other inputs, e.g., cameras 

%%%%%%%%%%%%%%%%%%%%%%%%%%%%%%%%%%%%%%%%%%%%%%%%%%%%%%%%%%%%%%%%%%%%%%%%%%%%%%%%
% \section{Proposed Method (Need to create a name)}
\section{Context-Aware DeLaN}
\label{sec:res_delan}
We will now briefly discuss DeLaN~\cite{delan_2021}, before presenting our extensions to the residual and contextual setting, as well as its application to online system identification for MPC.

DeLaN embeds Lagrangian mechanics into a deep learning framework to learn physically consistent dynamical models. 
%Consider the Lagrangian defined as the as $L = \EKin - \EPot$, where $\EKin = 1/2\jointVel\Transp\inertiaMat(\jointPos)\jointVel$ is the kinematic energy,
%
Consider the Lagrangian defined as the difference between the kinetic energy $\EKin = \frac{1}{2}\jointVel\Transp\inertiaMat(\jointPos)\jointVel$ and the potential energy $\EPot$. %From the Euler-Lagrange equation, an inverse dynamical model is obtained:
% \begin{equation}
%     \label{eq:delan_lagrangian}
%     \frac{d}{dt} \frac{\partial L}{\partial \jointVel} - \frac{\partial L}{\partial \jointPos} = \jointTau
% \end{equation}
Using the Euler-Lagrange equation, the system's equation of motion is given by
\begin{equation}
    \label{eq:inv_dyn}
    \inertiaMat(\jointPos) \jointAcc + \underbrace{\dot{\inertiaMat}(\jointPos) \jointVel
- \frac{1}{2} \left( \frac{\partial}{\partial \jointPos} 
\left( \jointVel\Transp \inertiaMat(\jointPos) \jointVel \right) \right)\Transp}_{:= \EoMC(\jointPos,\jointVel)\jointVel}
+ \underbrace{\frac{\partial \EPot}{\partial \jointPos}}_{:= \EoMG(\jointPos)} = \jointTau,
\end{equation}
where $\jointTau$ is the joint torque.

DeLaN approximates $\inertiaMat$ and $\EPot$ by two neural networks. The first one outputs the diagonal $\inertiaTriDiag$ and off-diagonal $\inertiaTriOff$ elements of the lower triangular matrix $\inertiaTri$ given by the Cholesky decomposition of $\inertiaMat$. Thus, the estimated inertia matrix $\hat{\inertiaMat}(\jointPos, \paramInertia)$ is given by
\begin{equation}
    \hat{\inertiaMat}(\jointPos, \paramInertia) = \hat{\inertiaTri}(\jointPos,\paramInertia)\hat{\inertiaTri}(\jointPos,\paramInertia)\Transp\text{,}
\end{equation}
where an offset $\epsilon$ and a softplus function is applied to $\inertiaTriDiag$ to guarantee $\hat{\inertiaMat}$ to be positive definite, and $\paramInertia$ is the network parameter. The second network directly outputs the potential energy $\hat{\EPot}(\jointPos, \paramPot)$. Therefore, the inverse dynamics model~\eqref{eq:inv_dyn} is defined as a function of the joint variables and the parameters of the networks $\jointTau = f^{-1}(\jointPos, \jointVel, \jointAcc, \paramInertia, \paramPot)$.

To train the network parameters, a loss function using inverse dynamics error can be employed. The gradient of the loss function is obtained using automatic differentiation in \eqref{eq:inv_dyn} of PyTorch \cite{pytorch_2024}.

\subsection{Residual DeLaN}
Given a nominal model, the equation of motion of the real system~\eqref{eq:inv_dyn} can be rewritten as
\begin{equation}
    \inertiaMatNom(\jointPos) \jointAcc + \EoMCNom(\jointPos,\jointVel) \jointVel + \EoMGNom(\jointPos) + \jointTauDiffNom = \jointTau\text{,}
\end{equation}
where $\inertiaMatNom$, $\EoMCNom$ and $\EoMGNom$ refer to the nominal components, and $\jointTauDiffNom=\jointTau - \jointTauNom$ denotes the residual torque between actual and nominal torque. 
Due to the model mismatches, i.e., $\inertiaMatDiff = \inertiaMat - \inertiaMatNom$, $\EoMGDiff = \EoMG - \EoMGNom$, the residual torque can be defined as
\begin{equation}
\label{eq:inv_dyn_diff}
    \jointTauDiffNom = \inertiaMatDiff(\jointPos) \jointAcc + \dot{\inertiaMatDiff}(\jointPos) \jointVel - \frac{1}{2} \left( \frac{\partial}{\partial \jointPos} \left( \jointVel\Transp \inertiaMatDiff(\jointPos) \jointVel \right) \right)\Transp + \EoMGDiff\text{,}
\end{equation}
which has the same form as \eqref{eq:inv_dyn} and can be learned using the DeLaN framework, i.e., $\jointTauDiffNom = \tilde{f}^{-1}(\jointPos, \jointVel, \jointAcc, \paramInertia, \paramPot)$.

Despite the extensive engineering efforts, model mismatches will always arise in real systems due to variations during production, flexible and moving parts, unmodelable internal components, e.g., wires and cables, and additional payloads attached to the robot. 
Hence, the residual torque is not only a function of joint states but also of the environment.

\subsection{Context-Aware DeLaN}
To enable DeLaN to estimate the residual torque $\jointTauDiffNom$ across different environments, we condition it on a latent embedding $\latentVariableZ$ of the environment. By training a history-based encoder along with the contextual DeLaN, we are able to infer the latent representation of the environment based on the recently observed states and actions. More specifically, we use an LSTM~\cite{hochreiter1997long} network as encoder, and feed it a sequence of joint positions, velocity and measured residual torques $\jointTauDiffNom$. Thus, each entry $i$ in the $\histLength$ length sequence is:
\begin{equation}
    \histInput_i = \begin{bmatrix}
        \jointPos_i\Transp &
        \jointVel_i\Transp &
        \jointTauDiffNom_i\Transp
    \end{bmatrix}\Transp\text{.}
\end{equation}

\begin{figure}[h]
    \centering
    \scalebox{1.4}{
    \includegraphics{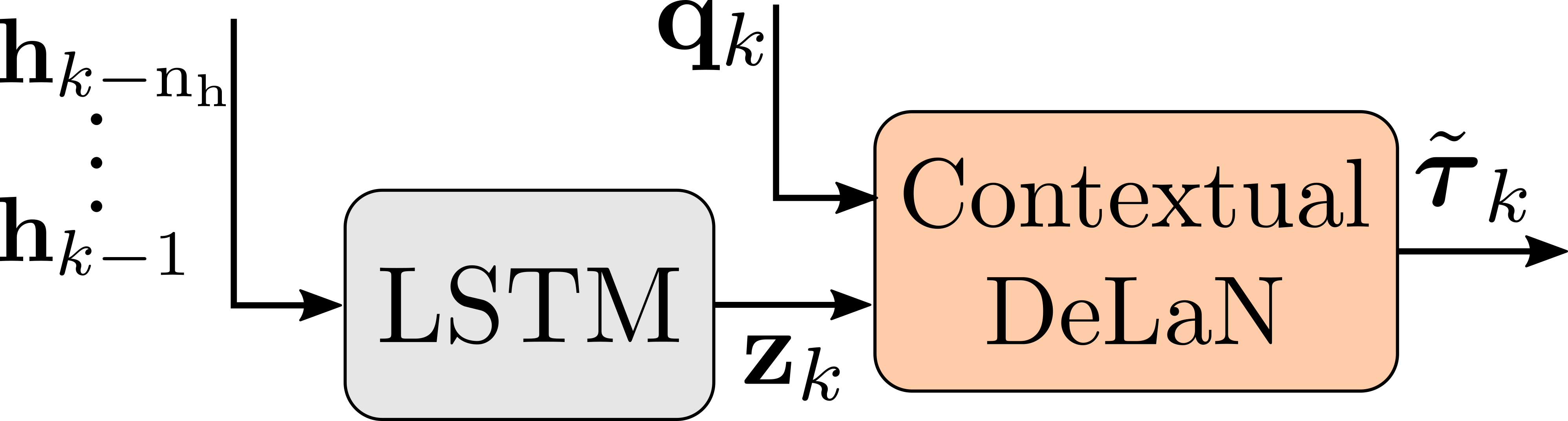}
}
    \caption{Our proposed architecture. The orange block is similar to the original DeLaN~\cite{delan_2021}, but with additional input $\latentVariableZ$.}
    \label{fig:residual_delan}
\end{figure}

We chose an LSTM for its ability to capture information from historical data while retaining long-term memory. By conditioning the contextual DeLaN on a learned latent representation, we can also reduce the number of neurons and layers. Please note, that we will only require the contextual DeLaN model during a given MPC iteration, but not the LSTM. The proposed architecture is depicted in Figure~\ref{fig:residual_delan}.
% \begin{figure}[h]
%     % \label{fig:residual_delan}
%     \centering
% \includegraphics[width=0.8\columnwidth]{Figures/contex_panda_delan_lstm_architecture_double.png}
%     \caption{Our proposed architecture. The orange block is similar to the original DeLaN~\cite{delan_2021}, but with additional input $\latentVariableZ$.}
%     \label{fig:residual_delan}
% \end{figure}
%

\subsection{Context-Aware DeLaN for MPC}
\label{sec:mpc}
We will now introduce CaDeLaC and discuss how it uses our context-aware and residual DeLaN model for MPC.

MPC is an optimal control technique that computes control actions in a receding horizon fashion by optimizing a cost function based on a system's model prediction~\cite{camacho_MPC}. To follow a desired state trajectory, we optimize at each instant $k$ the state and control inputs, $\Xopt_k$ and $\Uopt_k$, along the prediction horizon $\Ny$ with respect to the optimization problem
\begin{equation}
\begin{aligned}
    \label{eq:opt_mpc}
    &\underset{\Xopt_k,\Uopt_k}{\text{min}} &&
    \sum_{i=0}^{\Ny}\|{\xref_{k+i}} - \xss{}_{k+i}\|^2_{\Qmpc}  + \sum_{i=0}^{\Ny-1}\|\uss{}_{k+i}\|^2_{\Rmpc}\\
    &\textit{s.t.}&& \xss_k = \xss_0\text{,}
    \\&&&
 %   \label{eq:model_function}
    \xss_{k+i+1} = \fmodel(\xss{}_{k+i},\uss{}_{k+i})\text{,}
    \\&&&
 %   \label{eq:mpc_contraint}
    \hConst(\xss{}_{k+i}, \uss{}_{k+i}) \leq \ZeroMat\text{,}
\end{aligned}
\end{equation}
where $\Qmpc$ and $\Rmpc$ are the weights matrices, $\xss_0$ is the current state, $\xref_{k+i}$ is the desired state at each timestep ${k+i}$, $\fmodel(\xss{}_{k+i},\uss{}_{k+i})$ defines the system's model and $\hConst(\xss{}_{k+i}, \uss{}_{k+i})$ are inequality constraints, e.g., actions and state limits, collision avoidance or safety margins. 

A straightforward way to integrate our Context-Aware DeLaN would be to infer the residual torque along a future sequence of states and control actions, i.e., predicted in the step before or the reference values. This residual torque could then be given to the MPC as an external torque. However, the controller would only be able to compensate static model errors that do not depend on the system's state. 
Yet, we typically require state-dependent compensation for model mismatches, for example, due to mismatches in the inertia. A possible solution is to approximate the model using Taylor series \cite{gp_mpc, l4_casadi_taylor_salzmann2023neural}, but as we will discuss in more detail, we can also directly pass our model to the optimizer.

As we assume that the environment will not change within the prediction horizon, we can infer the latent variable $\latentVariableZ$ before every MPC iteration and use the resultant function $\tilde{f}^{-1}(\jointPos, \jointVel, \jointAcc)$ as a residual function. Please note, that the network parameters $\boldsymbol{\theta}=\{\paramInertia,\paramPot\}$ are trained jointly in an initial training phase, but remain constant throughout inference, as illustrated in Algorithm~\ref{alg:training} and Algorithm~\ref{alg:control}.

\begin{figure}[h]
\centering
\scalebox{1.4}{
\includegraphics{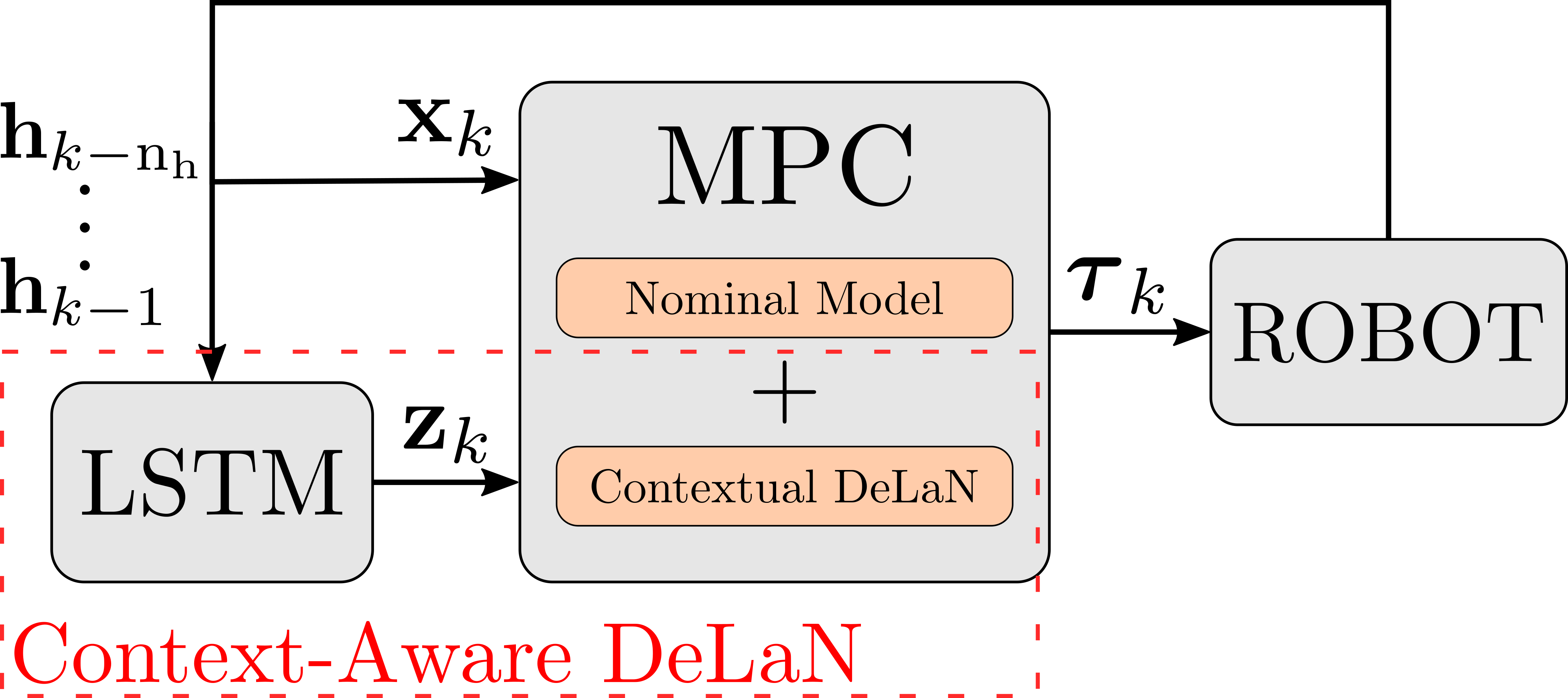}
}
\caption{CaDeLaC: Context-Aware Deep Lagrangian Model Predictive Control. The MPC optimizes considering both nominal and learned models (orange blocks).}
\label{fig:context_aware_mpc_architecture}
\end{figure}

\begin{algorithm}
\caption{Training Phase}
\begin{algorithmic}[1]
\REQUIRE Stepsize $\alpha$, Dataset $\mathcal{D} = \{\histInput_{1:T}^e\}_{e=1}^E$ of state-action sequences collected at different training environments $e$ (e.g., in simulation)
\FOR{every iteration $i$}
    \FOR{every environment $e \in \{1, \ldots, E\}$}
        \FOR{every timestep $t \in \{1, \ldots, T\}$}
            \STATE $\latentVariableZ \leftarrow \text{LSTM}(\histInput_0^e, \ldots, \histInput_t^e)$
            \STATE $\boldsymbol{\theta} \leftarrow \boldsymbol{\theta} - \alpha \nabla_{\boldsymbol{\theta}} \left\| \tilde{\jointTau}^e_t - f^{-1}(\jointPos_t^e, \jointVel_t^e, \jointAcc_t^e; \boldsymbol{\theta}, \latentVariableZ) \right\|^2$
        \ENDFOR
    \ENDFOR
\ENDFOR
\end{algorithmic}
\label{alg:training}
\end{algorithm}

\begin{algorithm}
\caption{Control Phase}
\begin{algorithmic}[1]
\REQUIRE Network parameters $\boldsymbol{\theta}$ obtained during training phase
\FOR{time step $k$}
    \STATE $\latentVariableZ_k \leftarrow \text{LSTM}(\histInput_{k-\mathrm{\histLength}}, \ldots, \histInput_{k-1})$
    \STATE $\uss_k \leftarrow$ solve (\ref{eq:opt_mpc}) with $f^{-1}(\jointPos_k, \jointVel_k, \jointAcc_k; \boldsymbol{\theta}, \latentVariableZ_k)$
    \STATE Apply control $\uss_k$
\ENDFOR
\end{algorithmic}
\label{alg:control}
\end{algorithm}

%%%%%%%%%%%%%%%%%%%%%%%%%%%%%%%%%%%%%%%%%%%%%%%%%%%%%%%%%%%%%%%%%%%%%%%%%%%%%%%%
%\input{03_control_architecture}

%%%%%%%%%%%%%%%%%%%%%%%%%%%%%%%%%%%%%%%%%%%%%%%%%%%%%%%%%%%%%%%%%%%%%%%%%%%%%%%%
\section{Experiments}
\label{sec:experiments}
In this section, we evaluate the proposed architecture to improve the joint trajectory tracking on a 7-DOF Franka Emika Panda robot with an unknown payload.

\subsection{Data Collection and Training}

We collect data using an LQR based on the nominal model for tracking a joint trajectory of the robot in simulation using MuJoCo~\cite{todorov2012mujoco}.  We collect 20 trajectories with a duration of 10 seconds for 100 different payloads attached to the end-effector. The mass and position of the payloads are uniformly sampled from 0~kg to 4~kg, while the 3D coordinates are sampled between -0.3~m and 0.3~m from the center of the end-effector. Additionally, we simulate the LQR without an additional load. 
The trajectories are sampled at the same control frequency of 50~Hz as the LQR; thus, the dataset has 1.01~M samples.
For system excitation, the joint reference trajectories are a chirp signal of 5th order and 10 seconds following \cite{friction_adaptive}, i.e.
\begin{gather}
    \label{eq:exc_reference}
    \begin{aligned}
    \jointPosRef(t) = \sum_{i=1}^{\chirpOrder} \frac{\chirpA{}_{i}}{i\chirpOmega} \sin(i\chirpOmega t) - \frac{\chirpB{}_{i}}{i\chirpOmega} \cos(i\chirpOmega t) + \jointPosInit\text{,}\\
    \jointPos_{\min} < \jointPosRef < \jointPos_{\max}, \quad \jointVel_{\min} < \jointVelRef < \jointVel_{\max}\text{,}
    \end{aligned}
\end{gather}
where $\chirpOmega = 2\pi/\chirpPeriod$ with $\chirpPeriod = 10$~s, $\chirpA$ and $\chirpB$ are uniformly sampled. The reference joint velocities $\jointVelRef$ are obtained by analytically differentiating the reference joint positions $\jointPosRef$.

In addition to the latent variable $\latentVariableZ$, a transformation input layer is applied to $\jointPos$ in both DeLaN networks, i.e., $\mathbf{T}_\jointPos (\jointPos) = [\cos(\jointPos), \sin(\jointPos)]$ as proposed in~\cite{delan_2021}. For the network architectures, we selected an MLP with 2 hidden layers, the first one with 30 neurons, and the second one with 20. %The LSTM has 5 hidden layers each with 10 neurons, with an output dimension of 10.
The LSTM has 5 hidden layers of 10 neurons each, followed by a dense layer applied to the last temporal output, yielding an output dimension of 10.

To enhance the robustness of our model to noise, we introduced artificial noise into the input dataset. The variances for the different joints and the training hyperparameters are detailed in the Appendix~A. %The model was trained for 3,000 epochs.

% Metrics for each joint in each environment:
% \begin{equation}
%     \text{dec} = \frac{1}{n_R}\sum_{r=0}^{n_R}\frac{\text{RMS}_{\text{nom}_{r}} - \text{RMS}_{\text{delan}_{r}}}{\text{RMS}_{\text{nom}_{r}}} 
% \end{equation}
% \begin{equation}
%     \text{dec_i} = \frac{1}{n_R}\sum_{r=0}^{n_R}\frac{\text{RMS}_{\text{nom}}_{ir}}{1}
    
%     % - \text{RMS}_{\text{delan}_{ir}}}{\text{RMS}_{\text{nom}_{ir}}} 
% \end{equation}

\subsection{MPC for Joint Trajectory Tracking}
We use the optimization problem presented in Section~\ref{sec:mpc} to track a joint trajectory, by choosing the state $\xss_k$ as the concatenation of joint velocities and positions, $\jointVel_k$ and $\jointPos_k$, while the actions $\uss_k$ correspond to the joint torques $\jointTau_k$. We use the inequality constraints $\mathbf{h}$ to bound the torques within their limits $\jointTau_\mathrm{min}$ and $\jointTau_\mathrm{max}$.

The nonlinear MPC is implemented using Acados~\cite{acados_2021} on SQP-RTI mode with the solver HPIPM~\cite{HPIPM}. The nominal model is obtained as a CasADi expression using Pinocchio~\cite{carpentier2019pinocchio}. To convert the PyTorch model to CasADI, we use the \textit{Naive} implementation of L4CasADi \cite{l4_casadi_taylor_salzmann2023neural} to convert the MLPs into symbolic expressions. 

While direct torque control is feasible in simulation, the control interface of the real robot requires a real-time loop controller at 1~kHz, which is challenging for any controller based on online optimization. Therefore, on the real robot, we use a feedback controller at 1~kHz with a feedforward torque that is updated by MPC at 50~Hz. 
The same frequencies are used for all evaluated controllers to ensure that performance differences are due only to the different models.
Hence, the commanded torque is computed as
\begin{equation}
    \jointTau_{\mathrm{CMD}} = \jointTau_{\mathrm{MPC}} + \PDKp(\jointPosRef - \jointPos) + \PDKd(\jointVelRef - \jointVel) - \EoMGNom(\jointPos)\text{,}
\end{equation}
where the feedback gains $\PDKp$ and $\PDKd$ are small, see Appendix~B, and only help to regularize the control action $\jointTau_{\mathrm{CMD}}$. 
To account for the robot's internal nominal gravity and friction compensations, we subtract $\EoMGNom(\jointPos)$ from $\jointTau_{\mathrm{CMD}}$.
Due to the presence of noise in the measured $\jointVel$ and the estimated $\jointAcc$ through finite differences, we apply a low-pass filter with a 2~Hz cutoff frequency to the LSTM's output.

\subsection{Extended Kalman Filter}
To show that only gravity compensation is insufficient for fast motions, an extended Kalman filter (EKF) is implemented to estimate the load as an external force $\Fee$ applied at the end-effector,
which is added to the nominal model,
\begin{equation}
    \inertiaMatNom(\jointPos)\jointAcc + \EoMCNom(\jointPos, \jointVel)\jointVel + \EoMGNom(\jointPos) = \jointTau + \Jee(\jointPos)\Transp\Fee\text{,}
\end{equation}
where $\Jee$ is the end-effector Jacobian. Similar to \cite{ext_torque_ekf}, we define the filter state and observation as
\begin{gather}
    \begin{aligned}
        \xss_a = \begin{bmatrix}
        \jointPos\Transp &
        \jointVel\Transp &
        \Fee\Transp
    \end{bmatrix}\Transp\text{,}
    & \ & 
    \yss_a = \begin{bmatrix}
        \jointPos\Transp &
        \jointVel\Transp
    \end{bmatrix}\Transp\text{,}
    \end{aligned}
\end{gather}
assuming a constant external force, $\dotFee = 0$. The estimated torque $\EKFTaueeEst = \Jee(\jointPos)\Transp\EKFFeeEst$ is provided to the MPC as an external torque applied to the joints, named EKF-MPC.

\subsection{Simulation - Experiments}
We evaluate our method in a wide range of environments. Namely, we evaluate Nominal MPC (using only the nominal model), EKF MPC, and CaDeLaC across 30 unseen environments over 20 trajectories, i.e., in total 600 trajectories of 10 seconds for each controller. The reference trajectories are given by \eqref{eq:exc_reference} with randomly sampled parameters. The payload parameters were sampled from the same range of mass and positions as for the training set.

To evaluate the performance of the three models, we analyze the Root Mean Square Error (RMSE) of the inferred residual torque. Table~\ref{tab:residual_torque_error_sim} presents the average RMSE of each model for all 1800 trajectories. 
The EKF reduced the residual torque error for only four of the seven joints. This occurs because the rotation axis of the first, along with the third and last joints in certain configurations, is parallel to gravity. Thus, the gravitational torque due to the load is zero, and only the extra inertia affects these joints, requiring some velocity or acceleration to manifest. Since the Context-Aware DelaN captures not only the gravitational component, it reduces all the errors significantly, outperforming the EKF. 
%
%Additionally, the rotation axis of the first joint (and the third joint in some configurations) is parallel to gravity. As a result, the gravitational torque is zero, and only the extra inertia significantly affects these joints, which requires some velocity or acceleration to manifest. 
\begin{table}[h]
    \centering
    \caption{Average residual torque RMSE $\bar{\tau}_\mathrm{e}$ for the three models over the 1800 simulated trajectories.}
    \setlength{\tabcolsep}{4pt} % Reducing column separation
    \small % Reducing font size
    \begin{tabular}{lccccccc}
        \toprule
        %Model & $\bar{\tau}_\mathrm{e\scriptstyle_1}$$\scriptstyle[\text{Nm}]$ & $\bar{\tau}_\mathrm{e\scriptstyle_2}$$\scriptstyle[\text{Nm}]$ & $\bar{\tau}_\mathrm{e\scriptstyle_3}$$\scriptstyle[\text{Nm}]$ & $\bar{\tau}_\mathrm{e\scriptstyle_4}$$\scriptstyle[\text{Nm}]$ & $\bar{\tau}_\mathrm{e\scriptstyle_5}$$\scriptstyle[\text{Nm}]$ & $\bar{\tau}_\mathrm{e\scriptstyle_6}$$\scriptstyle[\text{Nm}]$ & $\bar{\tau}_\mathrm{e\scriptstyle_7}$$\scriptstyle[\text{Nm}]$ \\
        \multirow{2}{*}{Model} & $\bar{\tau}_\mathrm{e\scriptstyle_1}$ & $\bar{\tau}_\mathrm{e\scriptstyle_2}$ & $\bar{\tau}_\mathrm{e\scriptstyle_3}$ & $\bar{\tau}_\mathrm{e\scriptstyle_4}$ & $\bar{\tau}_\mathrm{e\scriptstyle_5}$ & $\bar{\tau}_\mathrm{e\scriptstyle_6}$ & $\bar{\tau}_\mathrm{e\scriptstyle_7}$ \\
        & {\footnotesize$[\text{Nm}]$} &  {\footnotesize$[\text{Nm}]$} &  {\footnotesize$[\text{Nm}]$} &  {\footnotesize$[\text{Nm}]$} &  {\footnotesize$[\text{Nm}]$} &  {\footnotesize$[\text{Nm}]$} &  {\footnotesize$[\text{Nm}]$} \\
        \toprule
        \small Nominal & 2.28 & 10.55 & 5.13 & 11.63 & 3.21 & 3.98 & 2.56 \\
        Ours  & \textbf{0.95} & \textbf{1.36} & \textbf{1.81} & \textbf{1.37} & \textbf{0.92} & \textbf{0.81} & \textbf{0.83} \\
        EKF  & 10.02 & 8.40 & 11.30 & 7.25 & 1.87 & 1.25 & 2.56 \\
        \bottomrule
    \end{tabular}
    \label{tab:residual_torque_error_sim}
\end{table}

The average position and velocity tracking RMSE are presented in Table~\ref{tab:sim_traj_error}. CaDeLaC outperforms both Nominal MPC and EKF MPC in position tracking. However, for tracking velocity, the proposed controller presented larger errors in the first, second, and third joints than the nominal controller. 
One possible explanation is that the cost function considers both tracking errors and torque amplitudes, allowing for a different tradeoff due to the new controller's dynamic model.
The EKF MPC improved the tracking errors for the last three joints but significantly increased them for the first three joints due to the residual torque estimation and the trade-off within the cost function.
%As mentioned earlier when discussing our own performance, we attribute this to the trade-offs within the cost function.
\begin{table}[h]
    \centering
    \caption{Average tracking RMSE position $\bar{\mathbf{e}}$ and velocity $\bar{\dot{\mathbf{e}}}$ for the three controllers across 30 unseen simulated environments over 20 trajectories.}
    \setlength{\tabcolsep}{4pt}
    \begin{tabular}{lccccccc}
        \toprule
        \multirow{2}{*}{Controller} & $\bar{\jointPosErrNorm}_1$ & $\bar{\jointPosErrNorm}_2$ & $\bar{\jointPosErrNorm}_3$ & $\bar{\jointPosErrNorm}_4$ & $\bar{\jointPosErrNorm}_5$ & $\bar{\jointPosErrNorm}_6$ & $\bar{\jointPosErrNorm}_7$ \\
        &  $\footnotesize[\text{rad}]$ &  $\footnotesize[\text{rad}]$ &  $\footnotesize[\text{rad}]$ &  $\footnotesize[\text{rad}]$ &  $\footnotesize[\text{rad}]$ &  $\footnotesize[\text{rad}]$ &  $\footnotesize[\text{rad}]$ \\
        \toprule
        Nominal & 0.024  & 0.081  & 0.053  & 0.123  & 0.155  & 0.164  & 0.145  \\
        CaDeLaC & \textbf{0.022}  & \textbf{0.040}  & \textbf{0.025}  & \textbf{0.036}  & \textbf{0.037}  & 0.070  & \textbf{0.028}  \\
        EKF MPC & 0.093  & 0.083  & 0.114  & 0.081  & 0.101  & \textbf{0.066}  & 0.157  \\
        \midrule
        \multirow{2}{*}{Controller} & $\bar{\jointVelErrNorm}_1$ & $\bar{\jointVelErrNorm}_2$ & $\bar{\jointVelErrNorm}_3$ & $\bar{\jointVelErrNorm}_4$ & $\bar{\jointVelErrNorm}_5$ & $\bar{\jointVelErrNorm}_6$ & $\bar{\jointVelErrNorm}_7$ \\
        &  $\footnotesize[\text{rad/s}]$ &  $\footnotesize[\text{rad/s}]$ &  $\footnotesize[\text{rad/s}]$ &  $\footnotesize[\text{rad/s}]$ &  $\footnotesize[\text{rad/s}]$ &  $\footnotesize[\text{rad/s}]$ &  $\footnotesize[\text{rad/s}]$ \\
        \midrule
        Nominal & \textbf{0.059}  & \textbf{0.084}  & \textbf{0.096}  & \textbf{0.102}  & 0.212  & 0.215  & 0.221  \\
        CaDeLaC & 0.098  & 0.097  & 0.106  & 0.105  & \textbf{0.152}  & \textbf{0.163}  & \textbf{0.128}  \\
        EKF MPC & 0.168  & 0.152  & 0.192  & 0.166  & 0.201  & 0.176  & 0.267  \\
        \bottomrule
    \end{tabular}
    \label{tab:sim_traj_error}
\end{table}

\subsection{Hardware Experiments}

\subsubsection{High-speed trajectory}
\begin{figure}
     \centering
     \begin{subfigure}[b]{4cm}
         \centering
        \includegraphics[width=0.97\columnwidth]{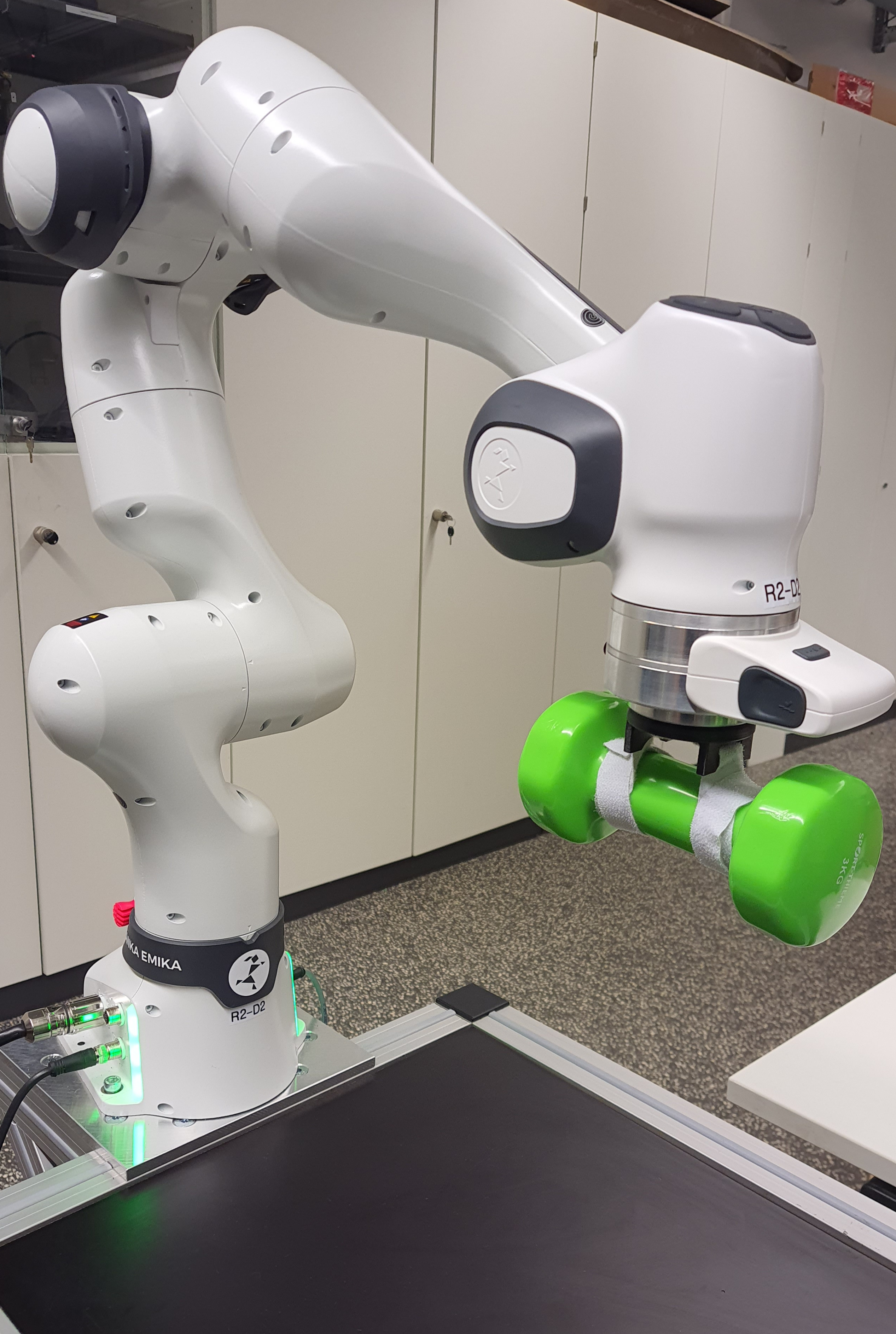}
        \caption{}
         \label{fig:robot_inf_symbol}
     \end{subfigure}
     \hfill
     \begin{subfigure}[b]{4cm}
         \centering
         \includegraphics[width=\columnwidth]{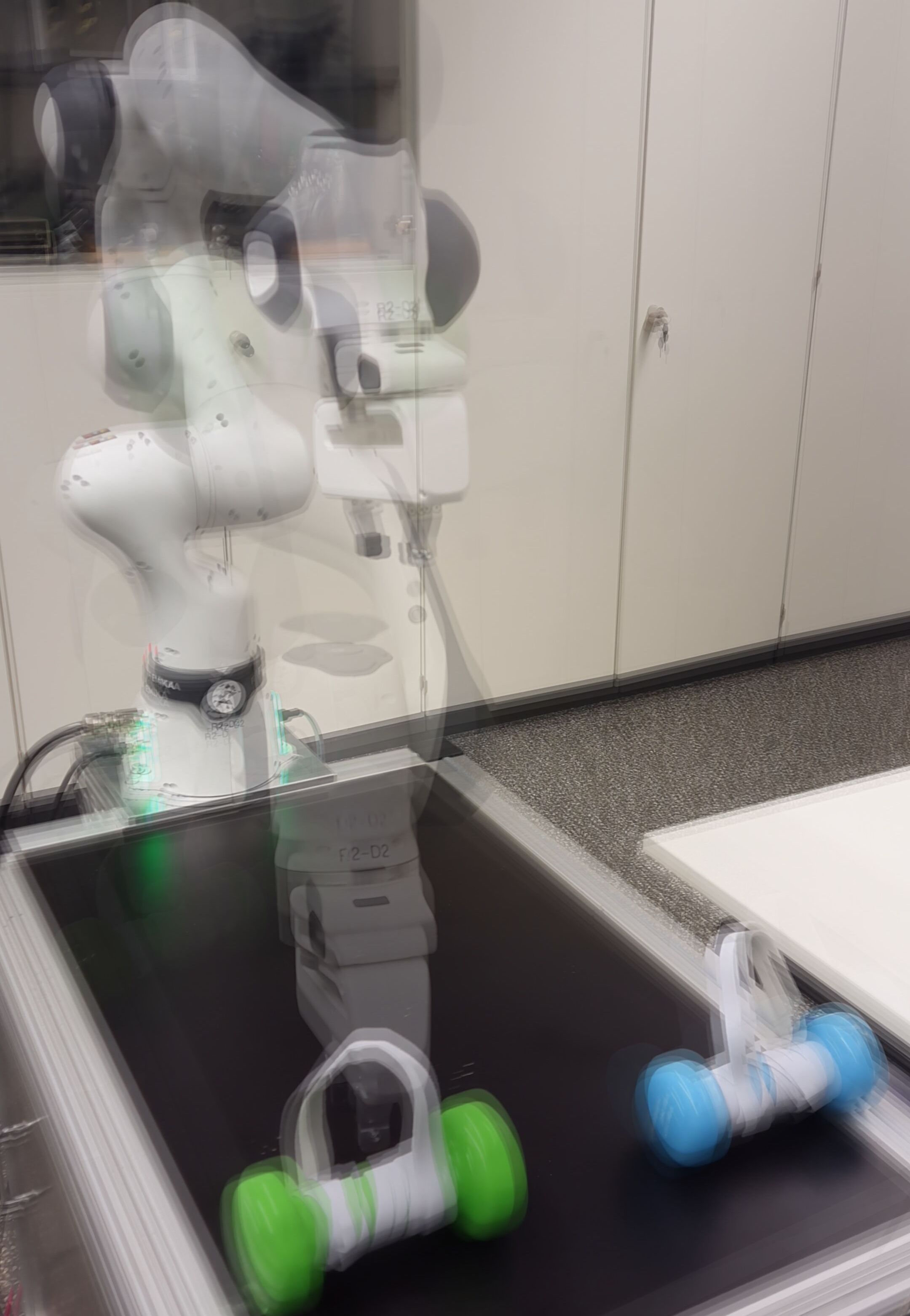}
         \caption{}
         \label{fig:robot_picknplace}
     \end{subfigure}
     \caption{Hardware experiments: (a) 3~kg gym weight attached to 3D-Printed holder; (b) The robot executes a sequence of two pick-and-place tasks with varying loads.}
     \label{fig:robot_setup}
\end{figure}

To evaluate our method in agile motions where gravity compensation alone might be insufficient, we tested all three controllers on high-speed trajectory tracking near the joint velocity limits with unseen payloads. 
Each controller was set to track the specified trajectory with three different gym weights: 1 kg, 2 kg, and 3 kg, resulting in a total of 9 trajectories. To ensure a secure attachment, we removed the robot's gripper and replaced it with a 3D-printed holder, depicted in Figure~\ref{fig:robot_inf_symbol}. Since the gripper weighs approximately 0.7~kg, the resulting payloads were 0.3~kg, 1.3~kg, and 2.3~kg, respectively. 

The high-speed trajectory is defined as:
\begin{equation}
\label{eq:inf_ref}
\pee{}_\mathrm{ref}(t) 
=
\begin{bmatrix}
0 \\
a_1 \sin(\omega_\mathrm{I} t) \cos(\omega_\mathrm{I} t) \\
a_2 \sin(\omega_\mathrm{I} t)
\end{bmatrix} + \pee{}_0\text{,}
\end{equation}
where $\pee{}_0$ is the initial end-effector position, $\omega_\mathrm{I} = 2 \pi f_\mathrm{I}$, with $f_\mathrm{I} = 0.7$~Hz, $a_1 = 0.40$, and $a_2 = 0.15$. The reference joint positions were obtained using inverse kinematics.
The reference joint velocities were then computed by finite difference, resulting in the following peak values: $\mathrm{max}(|\jointVel_\mathrm{ref}|) = [2.0, 2.22, 1.99, 1.3, 0.37, 0.95, 0.0])$. 

The inferred residual torque RMSE results are presented in Table~\ref{tab:exp_inf_residual_torque_rms}. Since the additional payload is only 0.3~kg for the 1~kg gym weight, the ratio between the residual torque and noise is small. As a result, both the EKF and our model do not improve significantly the residual torque.
Similar to the simulation results, the EKF was able to improve the residual torques for five of the seven joints, with no significant improvement in the first and third joints.
Our model presented an overall better estimation, as it also reduced the error for the first joint, while the errors for the last three joints remained small.
The residual torque results were not as good as in the simulation, mainly due to noise, friction, and an imperfect rigid attachment of the weights.
\begin{table}[h]
    \centering
    \caption{Average residual torque RMSE $\bar{\tau}_\mathrm{e}$ for the three models over the 9 high-speed trajectories.}
    \setlength{\tabcolsep}{4pt}
    \begin{tabular}{lccccccc}
        \toprule
        \multirow{2}{*}{Model} & $\bar{\tau}_\mathrm{e\scriptstyle_1}$ & $\bar{\tau}_\mathrm{e\scriptstyle_2}$ & $\bar{\tau}_\mathrm{e\scriptstyle_3}$ & $\bar{\tau}_\mathrm{e\scriptstyle_4}$ & $\bar{\tau}_\mathrm{e\scriptstyle_5}$ & $\bar{\tau}_\mathrm{e\scriptstyle_6}$ & $\bar{\tau}_\mathrm{e\scriptstyle_7}$ \\
        & {\footnotesize$[\text{Nm}]$} &  {\footnotesize$[\text{Nm}]$} &  {\footnotesize$[\text{Nm}]$} &  {\footnotesize$[\text{Nm}]$} &  {\footnotesize$[\text{Nm}]$} &  {\footnotesize$[\text{Nm}]$} &  {\footnotesize$[\text{Nm}]$} \\
        \midrule
        Nominal 1~kg & \textbf{3.03}  & 4.05  & \textbf{3.58}  & 2.58  & 0.47  & \textbf{0.70}  & \textbf{0.14}  \\
        Ours 1~kg & 3.95  & 4.12  & 4.27  & \textbf{2.47}  & 0.57  & 1.02  & 0.19  \\
        EKF 1~kg & 4.32  & \textbf{3.56}  & 4.83  & 2.67  & \textbf{0.43}  & 0.83  & \textbf{0.14}   \\
        \midrule
        Nominal 2~kg & 7.27  & 8.70  & 8.08  & 6.40  & \textbf{1.00}  & 1.55  & 0.42  \\
        Ours 2~kg & \textbf{6.36}  & 6.83  & \textbf{6.90}  & 4.45  & 1.24  & 1.86  & 0.53  \\
        EKF 2~kg & 7.26  & \textbf{6.04}  & 7.93  & \textbf{4.12}  & 1.01  & \textbf{1.47}  & \textbf{0.42}  \\
        \midrule
        Nominal 3~kg & 10.96  & 11.69  & 12.13  & 9.31  & 1.43  & 2.11  & \textbf{0.56}  \\
        Ours 3~kg & \textbf{9.40}  & \textbf{5.92}  & \textbf{9.68}  & 4.95  & 1.72  & 2.28  & 0.84  \\
        EKF 3~kg & 10.00  & 5.97  & 10.84  & \textbf{4.70}  & \textbf{1.37}  & \textbf{1.85}  & \textbf{0.56}  \\
        \bottomrule
    \end{tabular}
    \label{tab:exp_inf_residual_torque_rms}
\end{table}

The tracking errors are presented in Table~\ref{tab:exp_inf_tracking_error_rms}. 
Our controller outperformed position and velocity tracking for almost all the joints when compared to EKF MPC and Nominal MPC. Only the position tracking for $q_2$ and $q_6$ presented larger errors, probably due to trade-offs in the cost function, as mentioned before. As the last joint was the only one with a constant reference, its error analysis is irrelevant as they are very small. The EKF MPC still presented an overall performance than the nominal one, but it was not able to improve the errors for the first and third joints.
% Some values were larger compared to the ones of the nominal model ($ > 100\%$), probably due to trade-offs in the cost function as mentioned before. Regarding the second joint, our method presented an increase in the error for all the gym weights. This is a consequence of the trade-off, since joints 2 and 4 might conflict. 
% % 
% %Outperformed for the first four joints, while not so good for the last 2.
% Please note that the sixth-joint was the only one with a static reference, thus, its error analysis is irrelevant since the errors are very small.
\begin{table}[h]
    \centering
    \caption{Tracking position $\mathbf{e}$ and velocity $\dot{\mathbf{e}}$  RMSE for the three controllers over high-speed trajectories with three different loads.}
    \setlength{\tabcolsep}{3.7pt}
    \begin{tabular}{lccccccc}
        \toprule
        \multirow{2}{*}{Controller} & $\jointPosErrNorm_1$ & $\jointPosErrNorm_2$ & $\jointPosErrNorm_3$ & $\jointPosErrNorm_4$ & $\jointPosErrNorm_5$ & $\jointPosErrNorm_6$ & $\jointPosErrNorm_7$ \\
        &  $\footnotesize[\text{rad}]$ &  $\footnotesize[\text{rad}]$ &  $\footnotesize[\text{rad}]$ &  $\footnotesize[\text{rad}]$ &  $\footnotesize[\text{rad}]$ &  $\footnotesize[\text{rad}]$ &  $\footnotesize[\text{rad}]$ \\
        \toprule
        {\scriptsize Nominal 1~kg} & 0.032 & \textbf{0.020} & 0.024 & 0.054 & 0.017 & 0.015 & 0.002 \\
        {\scriptsize CaDeLaC 1~kg} & 0.032 & 0.034 & 0.028 & \textbf{0.036} & 0.009 & 0.019 & \textbf{0.001} \\
        {\scriptsize EKF MPC 1~kg} & \textbf{0.028} & 0.025 & \textbf{0.023} & 0.046 & \textbf{0.008} & \textbf{0.014} & 0.003 \\
        \midrule
        {\scriptsize Nominal 2~kg} & 0.044 & 0.027 & 0.035 & 0.091 & 0.016 & 0.026 & 0.007 \\
        {\scriptsize CaDeLaC 2~kg} & \textbf{0.029} & 0.052 & \textbf{0.020} & \textbf{0.036} & \textbf{0.011} & 0.022 & 0.005 \\
        {\scriptsize EKF MPC 2~kg} & 0.040 & \textbf{0.021} & 0.033 & 0.058 & \textbf{0.011} & \textbf{0.014} & \textbf{0.005} \\
        \midrule
        {\scriptsize Nominal 3~kg} & 0.053 & 0.057 & 0.045 & 0.124 & 0.022 & 0.031 & \textbf{0.002} \\
        {\scriptsize CaDeLaC 3~kg} & \textbf{0.033} & 0.076 & \textbf{0.028} & \textbf{0.031} & \textbf{0.014} & 0.027 & 0.009 \\
        {\scriptsize EKF MPC 3~kg}& 0.050 & \textbf{0.017} & 0.047 & 0.069 & 0.020 & \textbf{0.016} & 0.004 \\
        \midrule
        \multirow{2}{*}{Controller} & $\jointVelErrNorm_1$ & $\jointVelErrNorm_2$ & $\jointVelErrNorm_3$ & $\jointVelErrNorm_4$ & $\jointVelErrNorm_5$ & $\jointVelErrNorm_6$ & $\jointVelErrNorm_7$ \\
        &  $\footnotesize[\text{rad/s}]$ &  $\footnotesize[\text{rad/s}]$ &  $\footnotesize[\text{rad/s}]$ &  $\footnotesize[\text{rad/s}]$ &  $\footnotesize[\text{rad/s}]$ &  $\footnotesize[\text{rad/s}]$ &  $\footnotesize[\text{rad/s}]$ \\
        \midrule
        {\scriptsize Nominal 1~kg} & 0.379 & 0.243 & 0.317 & 0.150 & 0.100 & 0.133 & \textbf{0.008} \\
        {\scriptsize CaDeLaC 1~kg} & \textbf{0.326} & \textbf{0.196} & \textbf{0.261} & \textbf{0.128} & 0.105 & 0.162 & 0.015 \\
        {\scriptsize EKF MPC 1~kg}  & 0.368 & 0.221 & 0.322 & 0.166 & \textbf{0.081} & \textbf{0.122} & 0.009 \\
        \midrule
        {\scriptsize Nominal 2~kg} & 0.513 & 0.287 & 0.454 & 0.202 & \textbf{0.137} & 0.179 & 0.040 \\
        {\scriptsize CaDeLaC 2~kg} & \textbf{0.345} & \textbf{0.228} & \textbf{0.301} & \textbf{0.167} & \textbf{0.137 }& 0.255 & 0.052 \\
        {\scriptsize EKF MPC 2~kg} & 0.506 & 0.264 & 0.467 & 0.226 & 0.140 & \textbf{0.158} & \textbf{0.037} \\
        \midrule
        {\scriptsize Nominal 3~kg} & 0.603 & 0.293 & 0.554 & 0.254 & \textbf{0.201} & 0.236 & \textbf{0.035} \\
        {\scriptsize CaDeLaC 3~kg} & \textbf{0.420} & \textbf{0.260} & \textbf{0.405} & \textbf{0.213} & 0.232 & 0.311 & 0.124 \\
        {\scriptsize EKF MPC 3~kg} & 0.611 & 0.283 & 0.597 & 0.305 & 0.217 & \textbf{0.234} & 0.040 \\
        \bottomrule
    \end{tabular}
    \label{tab:exp_inf_tracking_error_rms}
\end{table}

\begin{figure*}[!t]
\centering
    \begin{subfigure}{.32\textwidth}
        \centering
        \includegraphics[width=0.99\textwidth]{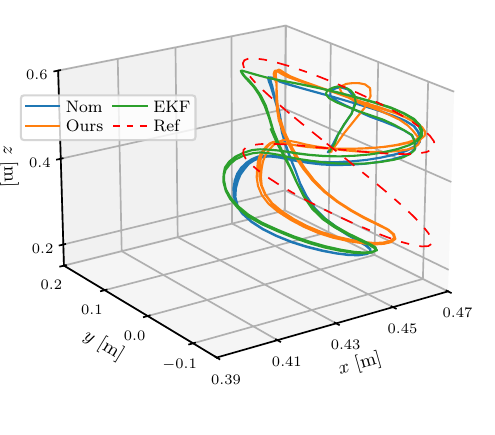}
        \vspace{-1.0cm}
        \caption{}
        \label{fig:exp_inf_1_kg}
    \end{subfigure}
    \begin{subfigure}{.32\textwidth}
        \centering
        \includegraphics[width=0.99\textwidth]{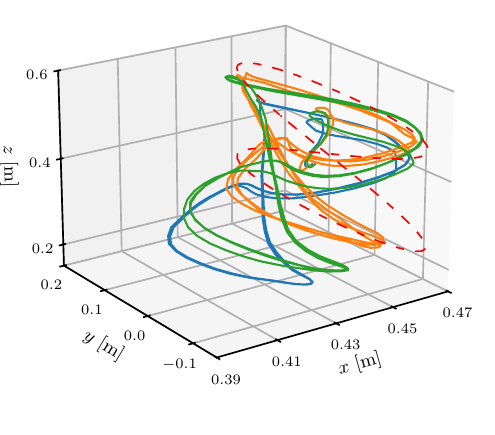}
        \vspace{-1.0cm}
        \caption{}
        \label{fig:exp_inf_2kg}
    \end{subfigure} 
    \begin{subfigure}{.32\textwidth}
        \centering
        \includegraphics[width=0.99\textwidth]{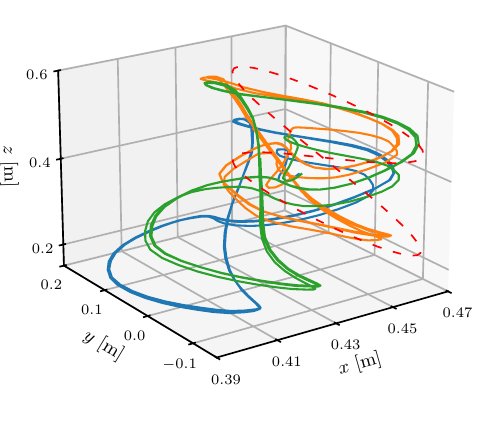}
        \vspace{-1.0cm}
        \caption{}
        \label{fig:exp_inf_3kg}
    \end{subfigure}
\caption{End-effector trajectory in high-speed experiments on the real robot. (a) 1~kg: CaDeLaC reduced the RMSE by 28.24\% (EKF MPC: 14.80\%); (b) 2~kg: 57.93\% (33.58\%); (c) 3~kg: 62.10\% (35.62\%).}
\label{fig:exp_inf_results}
\vspace{-5mm}
\end{figure*}

The first three cycles of the trajectories for each load, around 4.2 seconds, are presented in Figures~\ref{fig:exp_inf_1_kg}-\ref{fig:exp_inf_3kg}. Our method achieves the best tracking performance, reducing the RMSE in Cartesian space by 28.24\%, 57.93\%, and 62.10\% for 1~kg, 2~kg, and 3~kg, respectively, compared to 14.80\%, 33.58\%, and 35.62\% achieved by the EKF MPC.

The neural network, as part of the MPC's model, increases the optimization complexity and, consequently, the computational time. In Table~\ref{tab:exp_inf_times}, we present the average total time $\bar{t}_\mathrm{total}$ from the start of the MPC update to the moment the computed torque is sent, and the model time $\bar{t}_\mathrm{model}$ spent by the solver on model discretization and simulation. As the EKF does not add any complexity to the model, its times are comparable to the nominal one. CaDeLaC increased the total time by a factor of 4 due to the extra complexity of the model, as demonstrated by a higher $\bar{t}_\mathrm{model}$. However, the delay introduced by the additional computational time did not affect the hardware experiments, as they were almost 50\% below the control period of 20~ms. Also, due to the real-time kernel and the SQP-RTI, the total time variance was low, around $10^{-7}$.
\begin{table}[h]
    \centering
    \caption{Average total and model computational times in high-speed trajectories.}
    \setlength{\tabcolsep}{5pt} % Reducing column separation
    \begin{tabular}{lcc}
        \toprule
        Controller & $\bar{t}_\mathrm{total}$[ms] & $\bar{t}_\mathrm{model}$[ms]\\
        \toprule
        Nominal & \textbf{2.59} & 1.22 \\
        CaDeLaC & 10.57 & 8.50 \\
        EKF MPC & 2.70 & \textbf{1.19}  \\
        \bottomrule
    \end{tabular}
    \label{tab:exp_inf_times}
\end{table}

\subsubsection{Pick-and-place task}
% \begin{figure}
% \label{fig:picknplace}
% \centering
% \includegraphics[width=0.5 \linewidth]{Figures/Experiments/output_cut.jpg}
% \caption{The robot executes a sequence of two pick-and-place tasks with varying loads.}
% \end{figure}

To demonstrate that our method can adapt to dynamic changes in the environment, we define a \textit{pick-and-place} task, which consists of two repeated phases. In each phase, the robot must approach a gym weight, grasp it, perform a rapid motion in the air, and place it in a location different from the initial one.
%
%This task consists of two phases: the robot approach a gym weight, grasp it, perform a fast motion in the air, and place it in a location different from the initial one.
%
%Thus, the system's dynamics change four times along the trajectory.
During the trajectory, which lasts 26 seconds, the system's dynamics change four times. 
The robot waits 3 seconds at each pick-and-place position to ensure a safe grasp and disengagement of the strap. Thus, only 14 seconds of the full trajectory are spent in motion.
For this experiment, we reattached the gripper to the end-effector and attached straps to the gym load, since the gripper does not have sufficient force to directly hold the gym weight. The experimental setup is depicted in Figure~\ref{fig:robot_picknplace}. 
The slack caused by holding the strap introduced an additional challenge for the controller, as it allows the gym weight to swing, changing the relative position of the load's center of mass.
%
%We conducted experiments using 2 kg and 3 kg weights. 
The sequence of weights during the task was 3~kg and 2~kg.
Since the gripper was reattached, the additional loads relative to the nominal model are 2~kg and 3~kg, with the latter being the maximum payload specified by the manufacturer.

The results of the evaluated models are presented in Tables~\ref{tab:residual_torque_pp}. Similar to the simulation results and as previously mentioned, the EKF could not capture the residual torque for the first and third joints. While Context-Aware DeLaN showed significant improvement across all joints, except for the last joint, which already had a small error.
\begin{table}[h]
    \centering
    \caption{Average residual torque RMSE $\bar{\tau}_\mathrm{e}$ for the three models in the pick-and-place task.}
    \setlength{\tabcolsep}{4pt} % Reducing column separation
    \small % Reducing font size
    \begin{tabular}{lccccccc}
        \toprule
        \multirow{2}{*}{Model} & $\bar{\tau}_\mathrm{e\scriptstyle_1}$ & $\bar{\tau}_\mathrm{e\scriptstyle_2}$ & $\bar{\tau}_\mathrm{e\scriptstyle_3}$ & $\bar{\tau}_\mathrm{e\scriptstyle_4}$ & $\bar{\tau}_\mathrm{e\scriptstyle_5}$ & $\bar{\tau}_\mathrm{e\scriptstyle_6}$ & $\bar{\tau}_\mathrm{e\scriptstyle_7}$ \\
        & {\footnotesize$[\text{Nm}]$} &  {\footnotesize$[\text{Nm}]$} &  {\footnotesize$[\text{Nm}]$} &  {\footnotesize$[\text{Nm}]$} &  {\footnotesize$[\text{Nm}]$} &  {\footnotesize$[\text{Nm}]$} &  {\footnotesize$[\text{Nm}]$} \\
        \toprule
        \small Nominal & 0.69 & 7.97 & 0.78 & 6.74 & 0.19 & 1.93 & \textbf{0.03} \\
        Ours  & \textbf{0.64} & \textbf{2.32} & \textbf{0.59} & \textbf{1.55} & \textbf{0.14} & \textbf{0.45} & 0.04 \\
        EKF  & 1.70 & 2.78 & 1.65 & 2.21 & 0.24 & 0.63 & \textbf{0.03} \\
        \bottomrule
    \end{tabular}
    \label{tab:residual_torque_pp}
\end{table}

%
%The tracking errors are presented in Table~\ref{tab:pp_track_error}. Although our method had the best reduction for only 2 of 7 joints, it demonstrated an overall improvement while the EKF presented a large increase in a single joint, which degraded the end-effector tracking. The same is observed for the prediction error in Table~\ref{tab:pp_pred_error}. In this experiment we decided to analysis the prediction error because the trajectories are similar.

The tracking errors are presented in Table~\ref{tab:pp_track_error}. Although the EKF MPC had the best position tracking for $q_2$, $q_4$, and $q_5$, CaDeLaC presented an overall improvement in position and velocity tracking, particularly for $q_3$. Note that, since almost half of the full trajectory is spent in waiting positions, the RMSE values will not vary much among the controllers compared to a full trajectory in motion.
\begin{table}[h]
    \centering
    \caption{Tracking position $\mathbf{e}$ and velocity $\dot{\mathbf{e}}$  RMSE for the three controllers in the pick-and-place task.}
    \setlength{\tabcolsep}{4pt}
    \begin{tabular}{lccccccc}
        \toprule
        \multirow{2}{*}{Controller} & $\jointPosErrNorm_1$ & $\jointPosErrNorm_2$ & $\jointPosErrNorm_3$ & $\jointPosErrNorm_4$ & $\jointPosErrNorm_5$ & $\jointPosErrNorm_6$ & $\jointPosErrNorm_7$ \\
        &  $\footnotesize[\text{rad}]$ &  $\footnotesize[\text{rad}]$ &  $\footnotesize[\text{rad}]$ &  $\footnotesize[\text{rad}]$ &  $\footnotesize[\text{rad}]$ &  $\footnotesize[\text{rad}]$ &  $\footnotesize[\text{rad}]$ \\
        \toprule
        Nominal & \textbf{0.011} & 0.055 & 0.005 & 0.073 & 0.016 & 0.052 & \textbf{0.001} \\
        CaDeLaC   & \textbf{0.011} & 0.044 & \textbf{0.004} & 0.032 & 0.014 & \textbf{0.015} & \textbf{0.001} \\
        EKF MPC & \textbf{0.011} & \textbf{0.038} & 0.018 & \textbf{0.028} & \textbf{0.007} & \textbf{0.015} & \textbf{0.001} \\
        \midrule
        \multirow{2}{*}{Controller} & $\jointVelErrNorm_1$ & $\jointVelErrNorm_2$ & $\jointVelErrNorm_3$ & $\jointVelErrNorm_4$ & $\jointVelErrNorm_5$ & $\jointVelErrNorm_6$ & $\jointVelErrNorm_7$ \\
        &  $\footnotesize[\text{rad/s}]$ &  $\footnotesize[\text{rad/s}]$ &  $\footnotesize[\text{rad/s}]$ &  $\footnotesize[\text{rad/s}]$ &  $\footnotesize[\text{rad/s}]$ &  $\footnotesize[\text{rad/s}]$ &  $\footnotesize[\text{rad/s}]$ \\
        \midrule
        Nominal & \textbf{0.018} & 0.121 & 0.016 & 0.055 & 0.012 & 0.068 & \textbf{0.001} \\
        CaDeLaC & \textbf{0.018} & 0.104 & \textbf{0.014} & \textbf{0.049} & 0.011 & 0.061 & \textbf{0.001} \\
        EKF MPC & 0.019 & \textbf{0.101} & 0.023 & 0.057 & \textbf{0.008} & \textbf{0.048} & \textbf{0.001} \\
        \bottomrule
    \end{tabular}
    \label{tab:pp_track_error}
\end{table}

\begin{figure}%[t!]
\centering
    \includegraphics{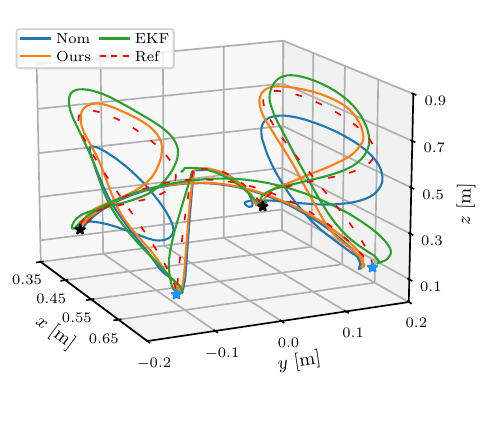}
    \vspace{-1.0cm}
    \caption{End-effector trajectory in the pick-and-place task for the three evaluated controllers. Blue and black markers indicate the pick and place positions, respectively. Each marker corresponds to a change in the payload and, consequently, in the system's dynamics.}
    \label{fig:traj_pick_and_place}
\end{figure}

Regarding the end-effector trajectory, the obtained trajectories for the three controllers are presented in Figure~\ref{fig:traj_pick_and_place}. Visually, CaDeLaC demonstrated better tracking, as it reduced the end-effector position RMSE by 39\%, compared to only 21\% with the EKF.

Since the controllers and hardware setup, including the computers and communication architecture, remain the same, we omit the computational time analysis, as it will be identical to the first hardware experiment.

%%%%%%%%%%%%%%%%%%%%%%%%%%%%%%%%%%%%%%%%%%%%%%%%%%%%%%%%%%%%%%%%%%%%%%%%%%%%%%%%
\section{Conclusion}
\label{sec:conclusion} In this work, we presented a method for online identification of physics-consistent DeLaN models and their application in model predictive control, named CaDeLaC. To enable online adaptation, we introduced latent environment embeddings as additional inputs to DeLaN and jointly learned a recurrent system identification network alongside contextual DeLaN. To meet the computational requirements of MPC, we proposed modeling only the residual dynamics with respect to a nominal model, enabling good predictive performance with small network sizes. Our real robot experiments demonstrated excellent tracking performance for fast motions under varying loads, outperforming the baselines. For future work, we want to apply our method to higher-dimensional and underactuated systems, such as humanoids. Furthermore, we want to investigate alternative objectives for training the history-based encoder, in particular, using reinforcement learning to directly optimize it with respect to downstream task performance.

%\addtolength{\textheight}{-1cm}   % This command serves to balance the column lengths

%%%%%%%%%%%%%%%%%%%%%%%%%%%%%%%%%%%%%%%%%%%%%%%%%%%%%%%%%%%%%%%%%%%%%%%%%%%%%%%%

%%%%%%%%%%%%%%%%%%%%%%%%%%%%%%%%%%%%%%%%%%%%%%%%%%%%%%%%%%%%%%%%%%%%%%%%%%%%%%%%

%%%%%%%%%%%%%%%%%%%%%%%%%%%%%%%%%%%%%%%%%%%%%%%%%%%%%%%%%%%%%%%%%%%%%%%%%%%%%%%%
% \section*{APPENDIX}False
\section*{APPENDIX}

\subsection{Data Collection and Training}
\label{appendix:data_collection}
The data collection and training are conducted on an AMD Ryzen 9 5900X processor with an NVIDIA GeForce RTX 4080 GPU and 32 GB of RAM. The parameters for both are presented in Tables~\ref{tab:data_parameters} and \ref{tab:train_parameters}.
\begin{table}[ht]
\centering
\caption{Data collection parameters.}
\setlength{\tabcolsep}{3pt}
\begin{tabular}{cc}
\toprule
\textbf{Parameter} & \textbf{Values} \\
\toprule
%
%$\Qlqr$, $\Rlqr$ & $\mathrm{diag}(100\times\mathbf{1}_{14})$, $\mathrm{diag}(0.05\times\mathbf{1}_{14})$ \\
%
$\Qlqr$ & $\mathrm{diag}(100\times\mathbf{1}_{14})$ \\
$\Rlqr$ & $\mathrm{diag}(0.05\times\mathbf{1}_{14})$ \\
$\jointPos_{\max}$ & {\scriptsize[2.8973, 1.7628, 2.8973, -0.0698, 2.8973, 3.7525, 2.8973]} \\
% \midrule
$\jointPos_{\min}$ & {\scriptsize[-2.8973, -1.7628, -2.8973, -3.0718, -2.8973, -0.0175, -2.8973]} \\
% \midrule
$\jointVel_{\max}$, $-\jointVel_{\min}$ & $2\times\mathbf{1}_{7}$ \\
% \midrule
%$\jointVel_{\min}$ & $-2\times\mathbf{1}_{7}$ \\
%
Range $\chirpA, \chirpB$ & $\pm$[0.20, 0.35, 0.58, 0.30, 0.58, 0.38, 0.58] \\
Samples & 1.01~M \\
\bottomrule
\end{tabular}
\label{tab:data_parameters}
\end{table}

\begin{table}[ht]
\centering
\caption{Training hyperparameters.}
\begin{tabular}{cc}
\toprule
\textbf{Parameter} & \textbf{Values} \\
\toprule
$\mathrm{var}(\jointVel)$ & $10^{-3}\times \mathbf{1}_{7}$ \\
$\mathrm{var}(\jointAcc)$ & [0.05, 0.05, 0.15, 0.03, 0.15, 0.25, 0.65] \\
$\mathrm{var}(\jointTau)$ & [0.5, 0.1, 0.5, 0.3, 0.01, 0.01, 0.01] \\
$\mathrm{var}(\jointTauDiffNom)$ & [1.0, 0.5, 1.0, 0.4, 0.02, 0.03, 0.02] \\
Activation & Tanh\\
Batch Size & 1024\\
Learning Rate & $5 \times 10^{-4}$\\
Epochs & 3000\\
%
%% Hidden layers already in the text
%DeLaN Hidden Layers & [30, 20]\\
%
%LSTM Hidden Layers & [10, 10, 10, 10, 10]\\
%
$\histLength$ & 15 \\
$\epsilon$ & 0.1 \\
\bottomrule
\end{tabular}
\label{tab:train_parameters}
\end{table}

\subsection{Experiments setup}
\label{appendix:exp_setup}
The inference and control experiments are performed on an AMD Ryzen 9 3900X processor with 64~GB of RAM, running Ubuntu 22.04 with a real-time kernel 5.15.0. All parameters related to the controllers are presented in Table~\ref{tab:control_parameters}.
\begin{table}[ht]
\centering
\caption{Controller parameters.}
\begin{tabular}{cc}
\toprule
\textbf{Parameter} & \textbf{Values} \\
\toprule
$\Ny$ & 12 \\
% \midrule
% Already in text
Control frequency & 50~Hz \\
% \midrule
Discretization time & 20~ms \\
% \midrule
$\Qmpc$ & $\mathrm{diag}(${\scriptsize$100 \times\mathbf{1}_{4}, 50 \times\mathbf{1}_{3}, 10^{3} \times \mathbf{1}_{4}, 500 \times\mathbf{1}_{3}$}$)$ \\
% \midrule
$\Rmpc$ & $\mathrm{diag}(0.2 \times\mathbf{1}_{4}, 1.0 \times \mathbf{1}_{3})$ \\
% \midrule
$\jointTau_\mathrm{max}$, $-\jointTau_\mathrm{min}$ & [87, 87, 87, 87, 12, 12, 12] \\
%
%$\jointTau_\mathrm{max}$, $\jointTau_\mathrm{min}$ & $\pm[87, 87, 87, 87, 12, 12, 12]$ \\
% \midrule
%$\jointTau_\mathrm{max}$ & -$\jointTau_\mathrm{max}$\\
% \midrule
%%
$\KFQ$ & $\mathrm{diag}(\mathbf{1}_{7}, 10^{-1} \times \mathbf{1}_{7}, 10^{2} \times\mathbf{1}_{3})$ \\
% \midrule
$\KFR$ & $\mathrm{diag}(10^{-2} \times \mathbf{1}_{14})$ \\
% \midrule
%%%
% \midrule
%$\Qlqr$ & $\mathrm{diag}(100\times\mathbf{1}_{14})$ \\
%
%$\Rlqr$ & $\mathrm{diag}(0.05\times\mathbf{1}_{14})$ \\
% \midrule
$\PDKp$ & [10, 10, 10, 10, 5, 5, 2] \\
% \midrule
$\PDKd$ & [2.5, 2.5, 2.5, 2.5, 1.5, 1.5, 1.5] \\
\bottomrule
\end{tabular}
\label{tab:control_parameters}
\end{table}

% \begin{equation}
% \label{eq:inv_dyn_diff}
%     (\inertiaMatNom + \inertiaMatDiff) \jointAcc + \dot{\inertiaMatDiff}(\jointPos) \jointVel
% - \frac{1}{2} \left( \frac{\partial}{\partial \jointPos} 
% \left( \jointVel\Transp \inertiaMatDiff(\jointPos) \jointVel \right) \right)\Transp
% + \EoMGDiff = \jointTauDiffNom \text{,}
% \end{equation}

% \begin{equation}
%     (\inertiaMatNom +  {\textcolor{red}\inertiaMatDiff})\jointAcc + (\EoMCNom + {\textcolor{red}\EoMCDiff}) \jointVel + \EoMGNom + {\textcolor{red}\EoMGDiff} = \jointTau\text{,}
% \end{equation}

% \begin{equation}
%     \inertiaMatDiff\jointAcc + \EoMCDiff \jointVel + \EoMGDiff = \jointTauDiffNom\text{,}
% \end{equation}


% \section*{ACKNOWLEDGMENT}

% This project has been funded by the German Federal Ministry of Research, Technology and Space (BMFTR) - Project number 01IS23057B.
% % The preferred spelling of the word ÒacknowledgmentÓ in America is without an ÒeÓ after the ÒgÓ. Avoid the stilted expression, ÒOne of us (R. B. G.) thanks . . .Ó  Instead, try ÒR. B. G. thanksÓ. Put sponsor acknowledgments in the unnumbered footnote on the first page.

%%%%%%%%%%%%%%%%%%%%%%%%%%%%%%%%%%%%%%%%%%%%%%%%%%%%%%%%%%%%%%%%%%%%%%%%%%%%%%%%

% \addtolength{\textheight}{-4.5cm}
% \addtolength{\textheight}{-0.45cm}
\addtolength{\textheight}{-1.1cm}
\bibliographystyle{./bibtex/IEEEtran}
\bibliography{main_root}

\begin{thebibliography}{10}
\providecommand{\url}[1]{#1}
\csname url@rmstyle\endcsname
\providecommand{\newblock}{\relax}
\providecommand{\bibinfo}[2]{#2}
\providecommand\BIBentrySTDinterwordspacing{\spaceskip=0pt\relax}
\providecommand\BIBentryALTinterwordstretchfactor{4}
\providecommand\BIBentryALTinterwordspacing{\spaceskip=\fontdimen2\font plus
\BIBentryALTinterwordstretchfactor\fontdimen3\font minus \fontdimen4\font\relax}
\providecommand\BIBforeignlanguage[2]{{%
\expandafter\ifx\csname l@#1\endcsname\relax
\typeout{** WARNING: IEEEtran.bst: No hyphenation pattern has been}%
\typeout{** loaded for the language `#1'. Using the pattern for}%
\typeout{** the default language instead.}%
\else
\language=\csname l@#1\endcsname
\fi
#2}}

\bibitem{slotine_regressor}
\BIBentryALTinterwordspacing
J.-J.~E. Slotine and W.~Li, ``On the adaptive control of robot manipulators,'' \emph{The International Journal of Robotics Research}, vol.~6, no.~3, pp. 49--59, 1987. [Online]. Available: \url{https://doi.org/10.1177/027836498700600303}
\BIBentrySTDinterwordspacing

\bibitem{est_inertia_an}
C.~H. An, C.~G. Atkeson, and J.~M. Hollerbach, ``Estimation of inertial parameters of rigid body links of manipulators,'' in \emph{1985 24th IEEE Conference on Decision and Control}, 1985, pp. 990--995.

\bibitem{friction_adaptive}
J.~Huang, D.~Tateo, P.~Liu, and J.~Peters, ``Adaptive control based friction estimation for tracking control of robot manipulators,'' \emph{IEEE Robotics and Automation Letters}, vol.~10, no.~3, pp. 2454--2461, 2025.

\bibitem{nadia}
J.~Foster, S.~McCrory, C.~DeBuys, S.~Bertrand, and R.~Griffin, ``Physically consistent online inertial adaptation for humanoid loco-manipulation,'' in \emph{2024 IEEE/RSJ International Conference on Intelligent Robots and Systems (IROS)}, 2024, pp. 11\,278--11\,285.

\bibitem{model_learning_control}
D.~Nguyen-Tuong and J.~Peters, ``Model learning for robot control: A survey,'' \emph{Cognitive processing}, vol.~12, pp. 319--40, 04 2011.

\bibitem{gpr_tuong}
M.~S. Duy Nguyen-Tuong and J.~Peters, ``Model learning with local gaussian process regression,'' \emph{Advanced Robotics}, vol.~23, no.~15, pp. 2015--2034, 2009.

\bibitem{gp_mpc}
L.~Hewing, J.~Kabzan, and M.~N. Zeilinger, ``Cautious model predictive control using gaussian process regression,'' \emph{IEEE Transactions on Control Systems Technology}, vol.~28, no.~6, pp. 2736--2743, 2020.

\bibitem{safe_gpr_mpc}
J.~Matschek, J.~Bethge, and R.~Findeisen, ``Safe machine-learning-supported model predictive force and motion control in robotics,'' \emph{IEEE Transactions on Control Systems Technology}, vol.~31, no.~6, pp. 2380--2392, 2023.

\bibitem{load_estimation_lwpr}
G.~Petkos and S.~Vijayakumar, ``Load estimation and control using learned dynamics models,'' in \emph{2007 IEEE/RSJ International Conference on Intelligent Robots and Systems}, 2007, pp. 1527--1532.

\bibitem{nmpc_rbf}
L.~F. Recalde, J.~Varela, B.~S. Guevara, V.~Andaluz, and D.~Gandolfo, ``Adaptive nmpc-rbf with application to manipulator robots,'' in \emph{2023 9th International Conference on Control, Decision and Information Technologies (CoDIT)}, 2023, pp. 2475--2482.

\bibitem{watson2024}
\BIBentryALTinterwordspacing
J.~Watson \emph{et~al.}, ``Machine learning with physics knowledge for prediction: A survey,'' 2024. [Online]. Available: \url{https://arxiv.org/abs/2408.09840}
\BIBentrySTDinterwordspacing

\bibitem{delan_lutter2018}
\BIBentryALTinterwordspacing
M.~Lutter, C.~Ritter, and J.~Peters, ``Deep lagrangian networks: Using physics as model prior for deep learning,'' in \emph{International Conference on Learning Representations}, 2019. [Online]. Available: \url{https://openreview.net/forum?id=BklHpjCqKm}
\BIBentrySTDinterwordspacing

\bibitem{delan_2021}
\BIBentryALTinterwordspacing
M.~Lutter and J.~Peters, ``Combining physics and deep learning to learn continuous-time dynamics models,'' \emph{CoRR}, vol. abs/2110.01894, 2021. [Online]. Available: \url{https://arxiv.org/abs/2110.01894}
\BIBentrySTDinterwordspacing

\bibitem{HNN}
S.~Greydanus, M.~Dzamba, and J.~Yosinski, ``Hamiltonian neural networks,'' in \emph{Advances in Neural Information Processing Systems}, vol.~32.\hskip 1em plus 0.5em minus 0.4em\relax Curran Associates, Inc., 2019.

\bibitem{delan_4ec}
\BIBentryALTinterwordspacing
M.~Lutter, K.~D. Listmann, and J.~Peters, ``Deep lagrangian networks for end-to-end learning of energy-based control for under-actuated systems,'' \emph{2019 IEEE/RSJ International Conference on Intelligent Robots and Systems (IROS)}, pp. 7718--7725, 2019. [Online]. Available: \url{https://api.semanticscholar.org/CorpusID:195874061}
\BIBentrySTDinterwordspacing

\bibitem{neural_ode_lie_control}
T.~Duong, A.~Altawaitan, J.~Stanley, and N.~Atanasov, ``Port-hamiltonian neural ode networks on lie groups for robot dynamics learning and control,'' \emph{IEEE Transactions on Robotics}, vol.~40, pp. 3695--3715, 2024.

\bibitem{model_knowledge_learn}
D.~Nguyen-Tuong and J.~Peters, ``Using model knowledge for learning inverse dynamics,'' in \emph{2010 IEEE International Conference on Robotics and Automation}, 2010, pp. 2677--2682.

\bibitem{residual_deep_residual_learning}
\BIBentryALTinterwordspacing
K.~He, X.~Zhang, S.~Ren, and J.~Sun, ``Deep residual learning for image recognition,'' \emph{2016 IEEE Conference on Computer Vision and Pattern Recognition (CVPR)}, pp. 770--778, 2015. [Online]. Available: \url{https://api.semanticscholar.org/CorpusID:206594692}
\BIBentrySTDinterwordspacing

\bibitem{meta_learning_nn_survey}
\BIBentryALTinterwordspacing
T.~Hospedales, A.~Antoniou, P.~Micaelli, and A.~Storkey, ``{Meta-Learning in Neural Networks: A Survey},'' \emph{IEEE Transactions on Pattern Analysis \& Machine Intelligence}, vol.~44, no.~09, pp. 5149--5169, Sept. 2022. [Online]. Available: \url{https://doi.ieeecomputersociety.org/10.1109/TPAMI.2021.3079209}
\BIBentrySTDinterwordspacing

\bibitem{gpr_hyperparamater}
\BIBentryALTinterwordspacing
R.~C. Grande, G.~V. Chowdhary, and J.~P. How, ``Nonparametric adaptive control using gaussian processes with online hyperparameter estimation,'' \emph{52nd IEEE Conference on Decision and Control}, pp. 861--867, 2013. [Online]. Available: \url{https://api.semanticscholar.org/CorpusID:7528974}
\BIBentrySTDinterwordspacing

\bibitem{Arcari2020MetaLM}
\BIBentryALTinterwordspacing
E.~Arcari, A.~Carron, and M.~N. Zeilinger, ``Meta learning mpc using finite-dimensional gaussian process approximations,'' \emph{ArXiv}, vol. abs/2008.05984, 2020. [Online]. Available: \url{https://api.semanticscholar.org/CorpusID:221112284}
\BIBentrySTDinterwordspacing

\bibitem{gpr_mpc_arm}
A.~Carron, E.~Arcari, M.~Wermelinger, L.~Hewing, M.~Hutter, and M.~N. Zeilinger, ``Data-driven model predictive control for trajectory tracking with a robotic arm,'' \emph{IEEE Robotics and Automation Letters}, vol.~4, no.~4, pp. 3758--3765, 2019.

\bibitem{Richards2021AdaptiveControlOrientedMF}
\BIBentryALTinterwordspacing
S.~M. Richards, N.~Azizan, J.-J.~E. Slotine, and M.~Pavone, ``Adaptive-control-oriented meta-learning for nonlinear systems,'' \emph{ArXiv}, vol. abs/2103.04490, 2021. [Online]. Available: \url{https://api.semanticscholar.org/CorpusID:232147745}
\BIBentrySTDinterwordspacing

\bibitem{bayesian_learning_mpc}
E.~Arcari, M.~V. Minniti, A.~Scampicchio, A.~Carron, F.~Farshidian, M.~Hutter, and M.~N. Zeilinger, ``Bayesian multi-task learning mpc for robotic mobile manipulation,'' \emph{IEEE Robotics and Automation Letters}, vol.~8, no.~6, pp. 3222--3229, 2023.

\bibitem{Lapandi2024MetaLearningAM_org}
\BIBentryALTinterwordspacing
D.~Lapandi{\'c}, F.~Xie, C.~K. Verginis, S.-J. Chung, D.~V. Dimarogonas, and B.~Wahlberg, ``Meta-learning augmented mpc for disturbance-aware motion planning and control of quadrotors,'' \emph{IEEE Control Systems Letters}, vol.~8, pp. 3045--3050, 2024. [Online]. Available: \url{https://api.semanticscholar.org/CorpusID:273229398}
\BIBentrySTDinterwordspacing

\bibitem{up_rss_2017}
\BIBentryALTinterwordspacing
W.~Yu, J.~Tan, C.~Karen~Liu, and G.~Turk, ``Preparing for the unknown: Learning a universal policy with online system identification,'' in \emph{Robotics: Science and Systems XIII}, ser. RSS2017.\hskip 1em plus 0.5em minus 0.4em\relax Robotics: Science and Systems Foundation, July 2017. [Online]. Available: \url{http://dx.doi.org/10.15607/RSS.2017.XIII.048}
\BIBentrySTDinterwordspacing

\bibitem{pinn}
\BIBentryALTinterwordspacing
M.~Raissi, P.~Perdikaris, and G.~Karniadakis, ``Physics-informed neural networks: A deep learning framework for solving forward and inverse problems involving nonlinear partial differential equations,'' \emph{Journal of Computational Physics}, vol. 378, pp. 686--707, 2019. [Online]. Available: \url{https://www.sciencedirect.com/science/article/pii/S0021999118307125}
\BIBentrySTDinterwordspacing

\bibitem{Li2023}
Q.~Li, T.~Wang, V.~Roychowdhury, and M.~K. Jawed, ``Metalearning generalizable dynamics from trajectories,'' \emph{Physical Review Letters}, vol. 131, no.~6, p. 067301, 2023.

\bibitem{knode_mpc}
K.~Y. Chee, T.~Z. Jiahao, and M.~A. Hsieh, ``Knode-mpc: A knowledge-based data-driven predictive control framework for aerial robots,'' \emph{IEEE Robotics and Automation Letters}, vol.~7, no.~2, pp. 2819--2826, 2022.

\bibitem{l4_casadi_taylor_salzmann2023neural}
T.~Salzmann, E.~Kaufmann, J.~Arrizabalaga, M.~Pavone, D.~Scaramuzza, and M.~Ryll, ``Real-time neural-mpc: Deep learning model predictive control for quadrotors and agile robotic platforms,'' \emph{IEEE Robotics and Automation Letters}, 2023.

\bibitem{Duong2021AdaptiveCO}
\BIBentryALTinterwordspacing
T.~P. Duong and N.~A. Atanasov, ``Adaptive control of se(3) hamiltonian dynamics with learned disturbance features,'' \emph{IEEE Control Systems Letters}, vol.~6, pp. 2773--2778, 2021. [Online]. Available: \url{https://api.semanticscholar.org/CorpusID:247597120}
\BIBentrySTDinterwordspacing

\bibitem{acados_2021}
R.~Verschueren \emph{et~al.}, ``acados -- a modular open-source framework for fast embedded optimal control,'' \emph{Mathematical Programming Computation}, 2021.

\bibitem{HPIPM}
G.~Frison and M.~Diehl, ``{HPIPM: a high-performance quadratic programming framework for model predictive control},'' \emph{IFAC-PapersOnLine}, vol.~53, no.~2, pp. 6563--6569, 2020, 21st IFAC World Congress.

\bibitem{carpentier2019pinocchio}
J.~Carpentier \emph{et~al.}, ``The pinocchio c++ library -- a fast and flexible implementation of rigid body dynamics algorithms and their analytical derivatives,'' in \emph{IEEE International Symposium on System Integrations (SII)}, 2019.

\bibitem{pytorch_2024}
\BIBentryALTinterwordspacing
J.~Ansel \emph{et~al.}, ``{PyTorch 2: Faster Machine Learning Through Dynamic Python Bytecode Transformation and Graph Compilation},'' in \emph{29th ACM International Conference on Architectural Support for Programming Languages and Operating Systems, Volume 2 (ASPLOS '24)}.\hskip 1em plus 0.5em minus 0.4em\relax ACM, Apr. 2024. [Online]. Available: \url{https://pytorch.org/assets/pytorch2-2.pdf}
\BIBentrySTDinterwordspacing

\bibitem{hochreiter1997long}
S.~Hochreiter and J.~Schmidhuber, ``Long short-term memory,'' \emph{Neural computation}, vol.~9, no.~8, pp. 1735--1780, 1997.

\bibitem{camacho_MPC}
E.~F. Camacho and C.~Bordons~Alba, \emph{Model predictive control}.\hskip 1em plus 0.5em minus 0.4em\relax Springer-Verlag London, 2007.

\bibitem{todorov2012mujoco}
E.~Todorov, T.~Erez, and Y.~Tassa, ``Mujoco: A physics engine for model-based control,'' in \emph{2012 IEEE/RSJ International Conference on Intelligent Robots and Systems}.\hskip 1em plus 0.5em minus 0.4em\relax IEEE, 2012, pp. 5026--5033.

\bibitem{ext_torque_ekf}
L.~Roveda, D.~Riva, G.~Bucca, and D.~Piga, ``External joint torques estimation for a position-controlled manipulator employing an extended kalman filter,'' in \emph{2021 18th International Conference on Ubiquitous Robots (UR)}, 2021, pp. 101--107.

\end{thebibliography}

\end{document}